\documentclass[10pt, a4paper]{article}

\usepackage[final]{lrec2026} % this is the new style
\usepackage{tgheros} % TeX Gyre Heros (Helvetica-like)
\usepackage{tgcursor}  % TeX Gyre Cursor (monospaced)
\usepackage{booktabs}

\usepackage{comment}
\usepackage{tabularx}
\usepackage[table]{xcolor} % for coloring rows
\usepackage{array}
\usepackage{caption}
\usepackage{subcaption}

\title{State of the Art in Text Classification for South Slavic Languages:\\Fine-Tuning or Prompting?}

\name{Taja Kuzman Pungeršek$^{\ast}$, Peter Rupnik$^{\ast}$, Ivan Porupski$^{\ast}$,
\\ {\bf \large Vuk Dinić$^{\ast}$, Nikola Ljubešić$^{\ast}$$^{\dagger}$$^{\ddagger}$ } } 

\address{$^{\ast}$Jožef Stefan Institute; \\ $^{\dagger}$Faculty of Computer and Information Science, University of Ljubljana; \\
$^{\ddagger}$Institute of Contemporary History;\\
Ljubljana, Slovenia\\
 \{taja.kuzman, peter.rupnik, ivan.porupski, vuk.dinic, nikola.ljubesic\}@ijs.si\\}

\abstract{
Until recently, fine-tuned BERT-like models provided state-of-the-art performance on text classification tasks. With the rise of instruction-tuned decoder-only models, commonly known as large language models (LLMs), the field has increasingly moved toward zero-shot and few-shot prompting. However, the performance of LLMs on text classification, particularly on less-resourced languages, remains under-explored. In this paper, we evaluate the performance of current language models on text classification tasks across several South Slavic languages. We compare openly available fine-tuned BERT-like models with a selection of open-weight and closed-source LLMs across three tasks in three domains: sentiment classification in parliamentary speeches, topic classification in news articles and parliamentary speeches, and genre identification in web texts. Our results show that LLMs demonstrate strong zero-shot performance, often matching or surpassing fine-tuned BERT-like models. Moreover, when used in a zero-shot setup, LLMs perform comparably in South Slavic languages and English. However, we also point out key drawbacks of LLMs, including less predictable outputs, significantly slower inference, and higher computational costs. Due to these limitations, fine-tuned BERT-like models remain a more practical choice for large-scale automatic text annotation.
 \\ \newline  }

\begin{document}

\maketitleabstract

\section{Introduction}

\renewcommand{\arraystretch}{1.2}

\begin{table*}[!ht]
\begin{center}
\begin{tabularx}{\textwidth}{|m{0.18\linewidth}|p{0.06\linewidth}|p{0.11\linewidth}|p{0.08\linewidth}|X|}
\hline
Dataset & Lang & \# Instances & \# Labels & \% Most and Least Frequent Label \\
\hline
\rowcolor{gray!20}
\multicolumn{5}{|c|}{\textit{Sentiment classification in parliamentary speeches}} \\
\hline
ParlaSent-EN-test & EN & 2600 & 3 & 40.8\% (Neutral), 26.8\% (Positive) \\
\hline
ParlaSent-HR-test & HR & 1336 & 3 & 41.9\% (Negative), 17.2\% (Positive) \\
\hline
ParlaSent-SR-test & SR & 1074 & 3 & 46.2\% (Negative), 17.6\% (Positive) \\
\hline
ParlaSent-BS-test & BS & 190 & 3 & 47.9\% (Negative), 14.7\% (Positive) \\
\hline
\rowcolor{gray!20}
\multicolumn{5}{|c|}{\textit{Genre classification in web texts}} \\
\hline
EN-GINCO & EN & 272 & 8 & 23.5\% (Information/Explanation), 0.4\% (Legal) \\
\hline
X-GINCO-SL & SL & 80 & 8 & 15\% (Prose/Lyrical), 8.8\% (Opinion/Argumentation) \\
\hline
X-GINCO-HR & HR & 80 & 8 & 16.3\% (Promotion), 7.5\% (Instruction) \\
\hline
X-GINCO-MK & MK & 80 & 8 & 15\% (News), 1\% (Opinion/Argumentation) \\
\hline
\rowcolor{gray!20}
\multicolumn{5}{|c|}{\textit{Topic classification in news articles}} \\
\hline
IPTC-test-HR & HR & 291 & 17 & 11.0\% (Economy), 3.8\% (Conflict, War and Peace) \\
\hline
IPTC-test-SL & SL & 282 & 17 & 10.6\% (Society), 3.2\% (Conflict, War and Peace) \\
\hline
\rowcolor{gray!20}
\multicolumn{5}{|c|}{\textit{Topic classification in parliamentary speeches}} \\
\hline
ParlaCAP-test-EN & EN & 876 & 22 & 6.4\% (Law and Crime), 2.1\% (Culture) \\
\hline
ParlaCAP-test-HR & HR & 869 & 22 & 8.5\% (Government Operations), 1.7\% (Immigration) \\
\hline
ParlaCAP-test-SR & SR & 874 & 22 & 7.1\% (Government Operations), 1.7\% (Immigration) \\
\hline
ParlaCAP-test-BS & BS & 824 & 22 & 10.4\% (Other), 0.5\% (Culture) \\
\hline
\end{tabularx}
\caption{Information on test datasets in English (EN), Croatian (HR), Serbian (SR), Bosnian (BS), Slovenian (SL), and Macedonian (MK).}
\label{tab:test-datasets}
    \end{center}
\end{table*}

Until recently, the dominant approach for text classification tasks relied on fine-tuning BERT-like transformer models on thousands of manually-annotated training examples. Recently, however, the field has shifted with the development of instruction-tuned decoder-only transformer models. These models, also commonly referred to as large language models (LLMs), which were originally developed primarily for text generation tasks, have demonstrated remarkable capabilities across a broad range of natural language processing (NLP) tasks, including text classification \citep{kuzman2023automatic,huang2023chatgpt}.

In this paper, we focus on South Slavic languages, where research on text classification tasks included in our study has, until recently, been limited or even non-existent \citep{kuzman2023survey, mochtak2024parlasent, kuzman-iptc-classification}. We take a first step toward systematically evaluating the current state of the art for text classification in these languages. Our evaluation is based on three text classification tasks in three different domains for which manually-annotated test datasets in South Slavic languages and fine-tuned BERT-like classifiers are freely available: sentiment classification of parliamentary speeches, topic classification in news articles, topic classification in parliamentary speeches, and automatic genre identification in web texts. These tasks span different domains and language styles, allowing for a comprehensive analysis of the performance of transformer-based models on text classification tasks. Specifically, we compare the performance of openly available fine-tuned BERT-like models with the zero-shot capabilities of both open-weight and closed-source LLMs used via prompting.

An important aspect of our study is to examine whether the performance of multilingual models on South Slavic languages is on par with their performance on English. This question is particularly relevant given that the evaluated large language models have been predominantly pretrained and instruction-tuned on English data.

By evaluating various models on a selection of text classification tasks in English and various South Slavic languages, we set out to test the following two hypotheses that are based on previous experiments with fine-tuned BERT-like models and LLMs on automatic genre identification \citep{kuzman2023automatic}, news topic classification \citep{kuzman-iptc-classification} and sentiment analysis in parliamentary texts \citep{mochtak2025parlasent}:
\begin{itemize}
    \item[H1] Zero-shot prompting with instruction-tuned large language models (LLMs) can achieve results comparable to the use of BERT-like models fine-tuned on training data that are similar to the test data.
    \item[H2] The performance of LLMs used in a zero-shot setup on text classification tasks on South Slavic test datasets is comparable to the performance on English test datasets.
\end{itemize}

%The evaluation comprises three text classification benchmarking datasets that include South Slavic languages and English: the BENCHić-sent Dataset for sentiment classification in parliamentary speeches \citep{mochtak2024parlasent,parlasent-repository}, X-GINCO and EN-GINCO datasets\footnote{\url{https://github.com/TajaKuzman/AGILE-Automatic-Genre-Identification-Benchmark}} for evaluation of automatic genre identification, and a IPTC News Topic test dataset \citep{kuzman-iptc-classification}. Further details on the datasets are provided in Section \ref{sec:datasets}. In the remainder of the paper, we provide information on the models included in the evaluation (Section \ref{sec:models}), and the results with discussion (Section \ref{sec:results}).

%Please use the \texttt{[review]} setting for submissions:

%\begin{verbatim}
%\usepackage[review]{lrec2026}
%\end{verbatim}

%This hides the authors and adds page numbers.

\section{Related Work}
\label{sec:related-work}

After the introduction of transformer architectures, BERT (bidirectional encoder representations from transformers) models have achieved state-of-the-art results in text classification tasks, outperforming earlier non-neural approaches, such as support vector machines (SVMs). They have also demonstrated strong cross-lingual zero-shot capabilities in various classification tasks, including automatic genre identification \citep{kuzman2023survey}, news topic classification \citep{petukhova2023mn,de2020news}, and sentiment classification \citep{mochtak2024parlasent}. However, these models still require fine-tuning on a training dataset, developed during manual annotation campaigns that are time-consuming and costly.

Instruction-tuned decoder-only transformer models, commonly referred to as large language models (LLMs), have recently shown strong performance in a range of classification tasks, even in zero-shot prompting setups that require no training data \citep{kuzman2023automatic,ljubevsic2024dialect,huang2023chatgpt,pungersek2026parlacap-paper}. They have achieved promising results on various natural language processing tasks, including stance detection \citep{zhang2022would}, implicit hate speech categorization \citep{huang2023chatgpt}, news topic classification \citep{kuzman-iptc-classification}, automatic genre identification \citep{kuzman2023automatic}, causal commonsense reasoning \citep{ljubevsic2024jsi}, and machine translation \citep{hendy2023good}. Due to their promising performance, researchers have even started using them as data annotators, either by generating text and labels \citep{meng2022generating} or by annotating pre-existing texts \citep{kuzman-iptc-classification,pungersek2026parlacap-paper}.
Despite the growing interest in this topic, the majority of evaluations of LLMs used in text classification tasks are limited only to English \citep{sun2023text,zhang2025pushing,kostina2025large,zhao2024advancing}. Systematic multilingual evaluations, especially which would include less-resourced languages such as those in the South Slavic group remain limited. Our work addresses this gap by providing a comparative evaluation of open-weight and closed-source LLMs with openly-available fine-tuned BERT-like models across four benchmark families comprising three diverse classification tasks and three different domains in South Slavic languages and English. %This study represents a first step toward benchmarking modern language models in this linguistic region, offering insights into their performance, limitations, and practical applicability.

\section{Benchmarks}
\label{sec:datasets}

The benchmarks (evaluation datasets) used in this study cover three text classification tasks, namely, sentiment identification, topic classification, and automatic genre identification, and three domains: parliamentary speeches, news articles and web texts. An overview of the datasets is provided in Table \ref{tab:test-datasets}. The four benchmark families differ significantly in terms of language coverage, number of test instances, and label granularity.

The topic classification task is evaluated on two domains: 1) news articles, namely, the Croatian and Slovenian IPTC test datasets \citep{kuzman-iptc-classification}, which comprise around 300 text instances per language, and 2) parliamentary speeches, namely, the Bosnian, Croatian, English and Serbian ParlaCAP test datasets \citep{pungersek2026parlacap-paper} that consist of approximately 820 to 880 instances per language. In the ParlaCAP benchmarks, an instance is a transcription of an utterance given by a parliamentary member in a parliamentary session.

The topic classification task involves the highest number of labels, that is, 17 news topic labels from the top level of the IPTC NewsCodes Media Topic hierarchical schema\footnote{\url{https://show.newscodes.org/index.html?newscodes=medtop&lang=en-GB&startTo=Show}} \citep{iptcGroupsNewsCodes}, and 22 policy topic labels (21 major topics and a label \textit{Other}) from the Comparative Agendas Project (CAP;  \citealp{baumgartner2019comparative}) Master Codebook~\citep{bevan2019gone}.\footnote{\url{https://www.comparativeagendas.net/pages/master-codebook}} 

In contrast, the Bosnian, Croatian, English, and Serbian ParlaSent sentiment identification datasets (\citealp{mochtak2024parlasent}; \citealp{parlasent-repository}) have a significantly lower granularity of labels, with only 3 categories. %However, this does not necessarily mean that the task is easier as sentiment might be harder to detect that the topic that can be identified based on more obvious lexical cues, such as keywords.
They are represented by the largest number of instances, ranging from 190 (Bosnian part) to 2600 (English part) sentence-level instances.

With 8 labels, the Croatian, English, Macedonian, and Slovenian GINCO genre datasets \citep{kuzman2023automatic} represent a midpoint in label granularity among the four benchmark families. However, the genre identification task might be the most difficult one, as genre identification depends on the interpretation of full texts with the focus on author's purpose, the common function of the text, and the text's conventional form \citep{orlikowski1994genre}. This complexity has also contributed to smaller test datasets in terms of the number of text instances, as manual annotation is more time-consuming. It is also important to note that, unlike the parliamentary datasets, the English portion of the genre datasets is not fully comparable to the South Slavic portions, which are label-balanced and contain fewer ambiguous instances. Nevertheless, the genre datasets remain valuable for evaluating model performance within each language. %, they are not suitable for cross-linguistic performance comparisons. 

All test datasets were manually annotated by annotators that are deemed reliable based on their satisfactory inter-annotator agreement, namely, Krippendorff’s alpha \citep{{krippendorff2018content}} values close to or above the 0.667 threshold for reliable annotation. To prevent large language models from incorporating the test datasets during their training phase, the test datasets are not publicly available, except for the ParlaSent benchmark family. Access to other datasets is granted on request from the corresponding authors. Further details on the test datasets are provided in Section \ref{sec:app-test-datasets} of the Appendix.

\section{Methodology}
\label{sec:models}

In this paper, we evaluate the main machine learning approaches that have recently been used for our selection of text classification tasks, with the focus on the comparison between the freely available fine-tuned BERT models and the open-weight and closed-source LLMs.\footnote{The code for the model evaluation and analysis of results is available at \url{https://github.com/TajaKuzman/Benchmarking-Text-Classification-on-South-Slavic}.}
The models are evaluated on four families of test datasets that comprise South Slavic languages. The performance of the models is evaluated based on the micro-F1 and macro-F1 metrics, which enable assessment of the model performance at both the instance and label levels, respectively.

The following machine learning models are included in the evaluation:
\begin{itemize}
    \item \textbf{dummy classifier}: a dummy classifier that predicts the most frequent class in the training data. To allow comparison, the dummy classifiers were trained on the same datasets that were used for fine-tuning the BERT-like models, mentioned below.
    %\item \textbf{non-neural baseline classifiers}: a dummy classifier that predicts the most frequent class in the training data; and two non-neural machine learning methods that have been frequently used for text classification tasks in the past, namely the Naive Bayes classifier and support vector machines (SVMs). To allow comparison, the classifiers were trained on the same datasets that were used for fine-tuning the BERT-like models, mentioned below.
    \item \textbf{fine-tuned BERT-like classifiers}: in our study, we evaluate previously developed openly accessible multilingual fine-tuned BERT-like models that have been fine-tuned for the respective task, namely, the XLM-R-ParlaSent (\citealp{parlasent-model}; \citealp{mochtak2024parlasent}) model for sentiment identification in parliamentary texts, the X-GENRE classifier (\citealp{kuzman2023automatic}; \citealp{x-genre-repository,x-genre-classifier-huggingface}) for automatic genre identification, the IPTC News Topic classifier (\citealp{kuzman-iptc-classification}; \citealp{iptc_model_huggingface}) for news topic classification, and the ParlaCAP classifier (\citealp{pungersek2026parlacap-paper}; \citealp{parlacap_model}) for topic classification in parliamentary speeches. The XLM-R-ParlaSent and the ParlaCAP models are based on the XLM-R-parla pretrained model \citep{xlm-r-parla} that was developed by additionally pretraining the large-sized XLM-RoBERTa model \citep{conneau2020unsupervised} on parliamentary proceedings in 30 European languages \citep{mochtak2024parlasent}. The XLM-R-ParlaSent model was fine-tuned on 13 thousand instances from the ParlaSent sentiment training dataset \citep{parlasent-repository} in seven European languages (Bosnian, Croatian, Czech, English, Serbian, Slovak, and Slovenian; \citealp{mochtak2024parlasent}), while the ParlaCAP model was fine-tuned on the ParlaCAP-train dataset (\citealp{parlacap-train}; \citealp{pungersek2026parlacap-paper}) that comprises around 30 thousand speeches from parliamentary debates annotated with CAP topic labels, originating from the ParlaMint 4.1 parliamentary datasets (\citealp{parlamint_41}; \citealp{erjavec2025parlamint}) in 29 European languages. The X-GENRE classifier is based on the base-sized XLM-RoBERTa model \citep{conneau2020unsupervised} and was fine-tuned on the training split of the X-GENRE dataset \citep{x-genre-dataset} in English and Slovenian; while the IPTC News Topic classifier is based on the large-sized XLM-RoBERTa model \citep{conneau2020unsupervised} that was fine-tuned on the EMMediaTopic dataset \citep{emmediatopic} in Catalan, Croatian, Greek, and Slovenian. All fine-tuned models use the same classes as the test datasets used in our study.
    \item \textbf{open-weight and closed-source large language models}: we use closed-source OpenAI models, namely the GPT-3.5-Turbo (\texttt{gpt-3.5-turbo-0125}; \citealp{openai}), GPT-4o (\texttt{gpt-4o-2024-08-06}; \citealp{openai-gpt4o}) and the GPT-5 (\texttt{gpt-5-2025-08-07}; \citealp{gpt-5}); a closed-source Gemini 2.5 Flash model \citep{comanici2025gemini} by Google DeepMind; a closed-source Mistral Medium 3.1 model (\texttt{mistral-medium-2508}; \citealp{mistral}) by Mistral AI; and four open-weight models, namely, the Meta LLaMA 3.3 model \citep{llama33modelcard}, the Gemma 3 model \citep{gemma3}, the Qwen 3 model \citep{yang2025qwen3}, and the DeepSeek-R1-Distill model (\texttt{DeepSeek-R1-Distill-Qwen-14B}; \citealp{guo2025deepseek}). It is important to note that while the LLaMA model was pretrained on a web text collection in various languages, it is said to support only 8 languages, namely English, German, French, Italian, Portuguese, Hindi, Spanish, and Thai \citep{llama33modelcard}. The DeepSeek-R1-Distill model is based on the Qwen 2.5 model \citep{qwen2.5,team2024qwen2} that provides support for more than 29 languages -- not including South Slavic languages though. In contrast, the Gemma 3 model is reported to support over 140 languages \citep{gemma3}, and the Qwen 3 model was pretrained on 119 languages \citep{yang2025qwen3}. While closed-source models are said to be massively multilingual, with Gemini 2.5 models being pretrained on over 400 languages \citep{comanici2025gemini}, details on their language coverage are very limited.
\end{itemize}

Open-weight models were installed locally and executed via the Ollama API service \citep{maric_local_llms}. OpenAI models were used through the chat completion endpoint via the OpenAI API, whereas other closed-source models were accessed through the OpenRouter platform\footnote{\url{https://openrouter.ai/}} that provides a unified API access to various closed-source models. To prevent any bias, all models were used with their default parameters. The only parameter that we defined is the temperature which we set to 0 to ensure a more deterministic behaviour of the models. More details on the models and their implementation, including information on the availability of openly available models and fine-tuning datasets, are provided in Section \ref{app:sec-models} of the Appendix.

All instruction-tuned LLMs are used in a zero-shot prompting setup, meaning that they receive only a task description and label definitions. 
All prompts and label definitions are written in English, while the instances are provided in the original languages (English or South Slavic). Changing the language of the prompt could introduce an additional factor that affects performance. In the presented experiments, our goal is to assess model performance on a specific task and language, rather than to evaluate their instruction-following abilities across different languages.  
The models are instructed to output a label, represented by a digit. The same prompt per benchmark family is used for all LLMs. %In rare cases, models returned invalid outputs (e.g., \texttt{\{"genre": -1\}}, \texttt{\{"genre": 13\}}). These were corrected to the \textit{Mix} label, which is also used by the X-GENRE classifier to indicate low prediction confidence.
Prompts are provided in Figure \ref{fig:prompts} in Section \ref{app:sec-models} of the Appendix.

\section{Results}
\label{sec:results}

\begin{figure*}[htbp]
    \centering

    \begin{subfigure}[b]{0.45\textwidth}
        \centering
        \includegraphics[width=\textwidth]{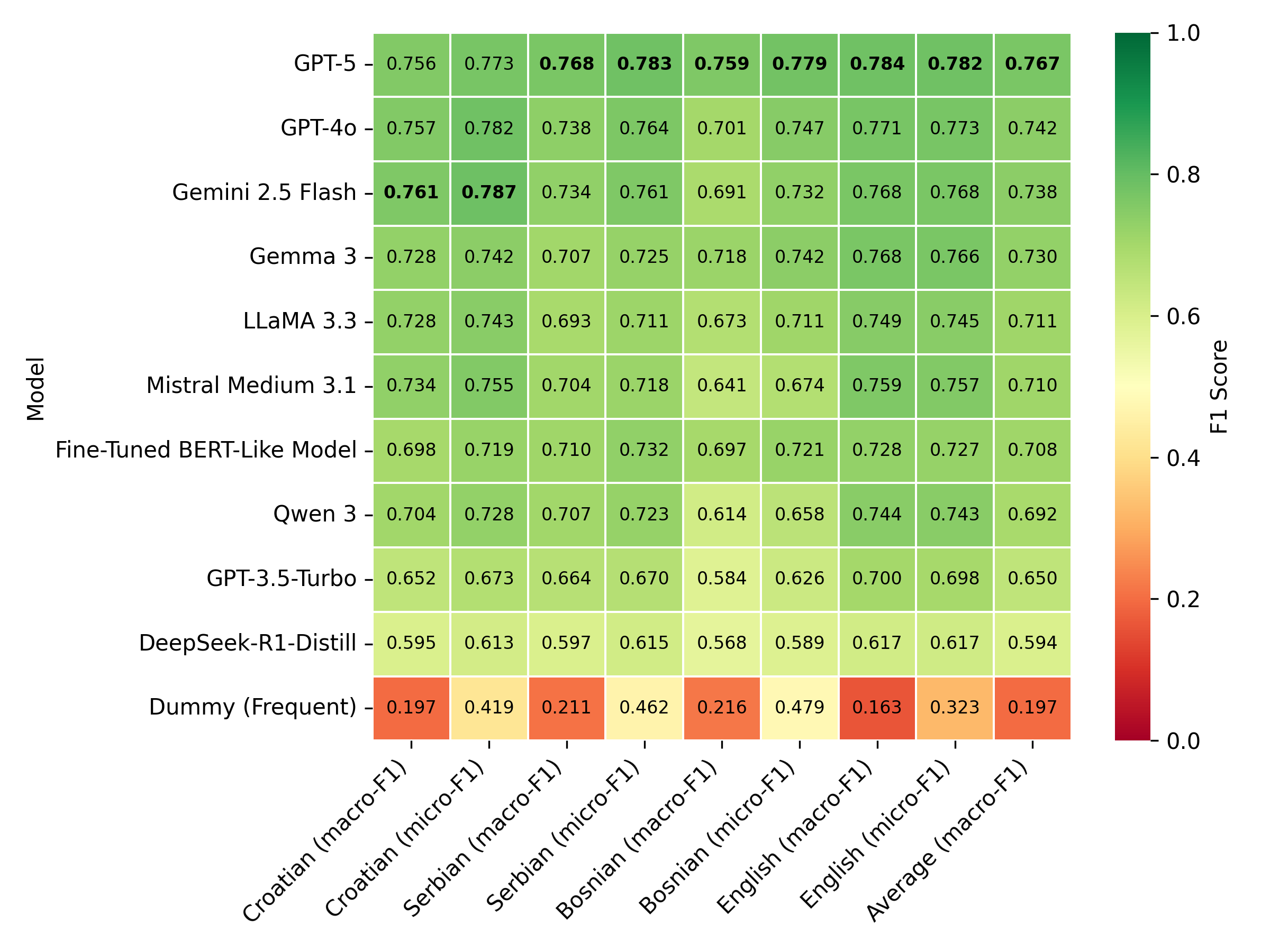}
        \caption{Sentiment classification.}
        \label{fig:sentiment-results}
    \end{subfigure}
    \hfill
    \begin{subfigure}[b]{0.45\textwidth}
        \centering
        \includegraphics[width=\textwidth]{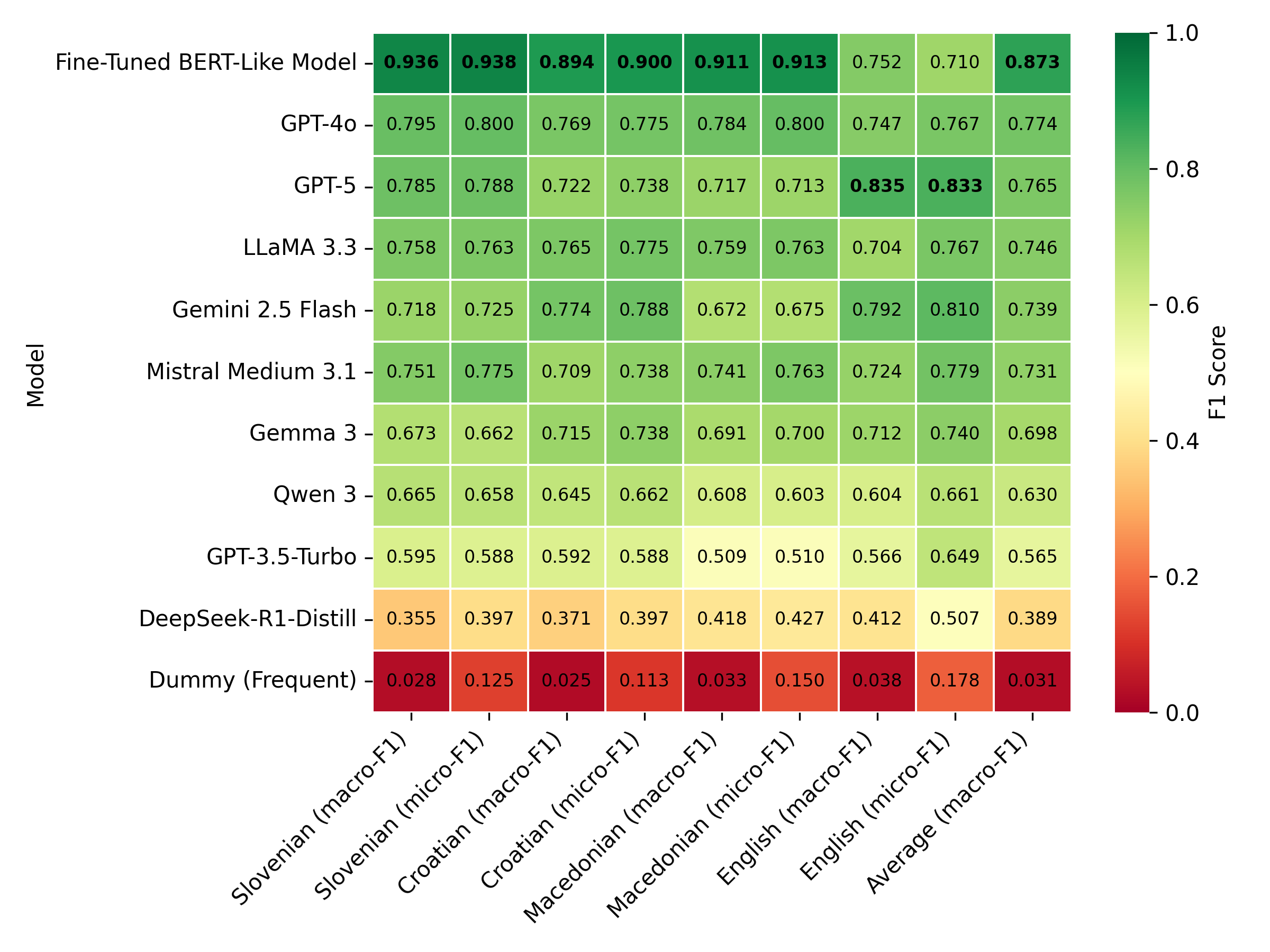}
        \caption{Automatic genre identification.}
        \label{fig:genre-results}
    \end{subfigure}

    \vskip\baselineskip 

    % Second row
    \begin{subfigure}[b]{0.45\textwidth}
        \centering
        \includegraphics[width=\textwidth]{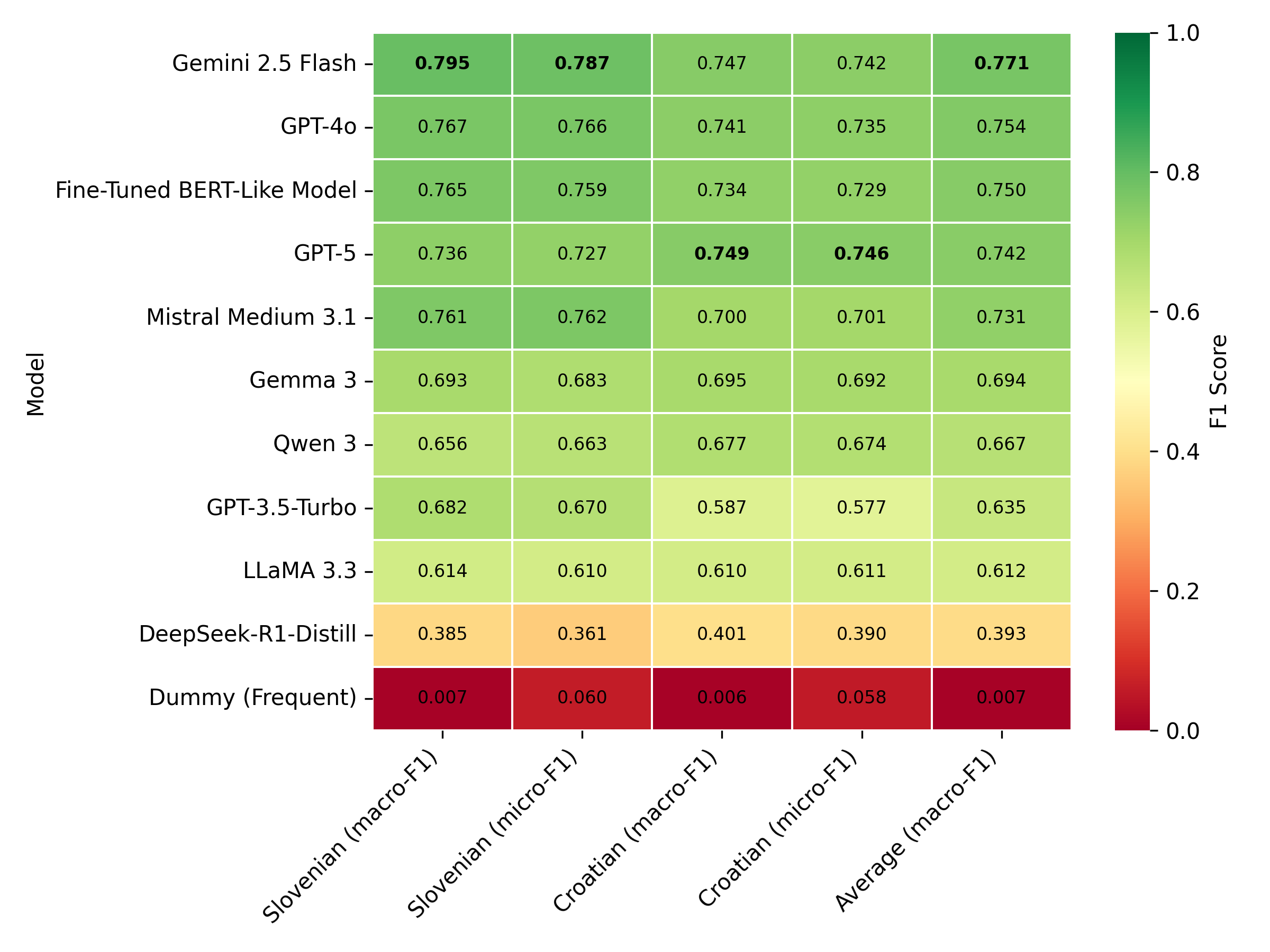}
        \caption{News topic classification.}
        \label{fig:iptc-results}
    \end{subfigure}
    \hfill
    \begin{subfigure}[b]{0.45\textwidth}
        \centering
        \includegraphics[width=\textwidth]{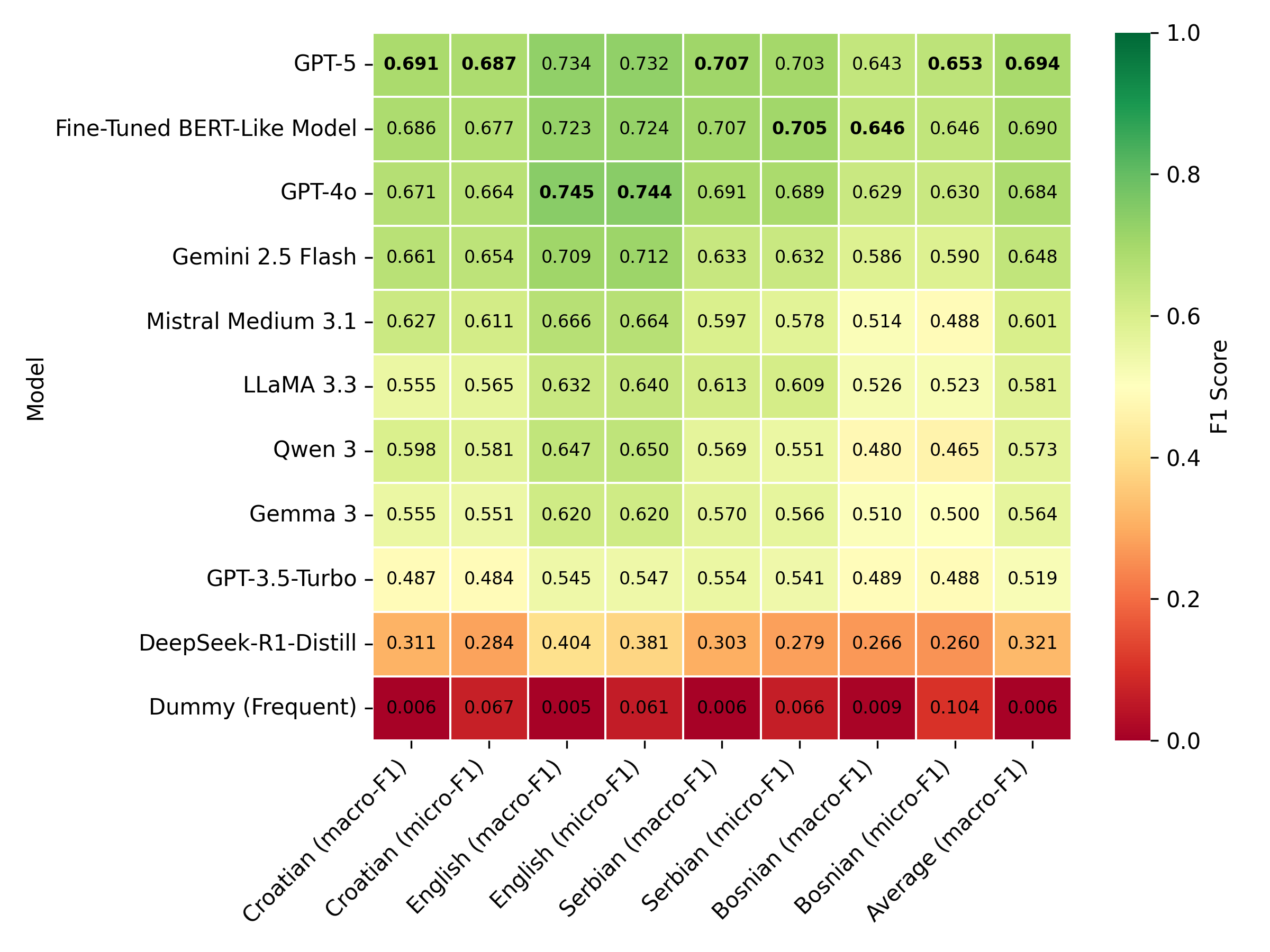}
        \caption{Parliamentary topic classification.}
        \label{fig:parlacap-results}
    \end{subfigure}
    
    \caption{Micro-F1 and macro-F1 scores across models and languages on the test datasets for sentiment classification (Figure \ref{fig:sentiment-results}), automatic genre identification (Figure \ref{fig:genre-results}), and topic classification on news (Figure \ref{fig:iptc-results}) and parliamentary speeches (Figure \ref{fig:parlacap-results}).}
    \label{fig:main-results}
\end{figure*}

In this section, we evaluate the performance of the fine-tuned BERT-like models and the instruction-tuned LLMs on a selection of text classification tasks that include test datasets in South Slavic languages. First, in Section \ref{sec:sota-results}, we provide results on the four benchmark families with the focus on hypothesis H1, which expects that zero-shot prompting with LLMs can provide performance that is comparable to that of fine-tuned BERT-like models. In Section \ref{sec:gpt-comparison}, we compare in more detail the performance of the closed-source and open-weight LLMs on the three text classification tasks, which is followed by a discussion on the advantages and limitations of LLMs for data annotation based on text classification tasks (Section \ref{sec:advantages-llms}). Lastly, in Section \ref{sec:results-en-vs-south-slavic}, we compare the performance of LLMs on English test datasets with their performance on South Slavic datasets, addressing hypothesis H2, which presumes that the available multilingual LLMs perform similarly on South Slavic languages as on English.

\begin{table}[!ht]
    \centering
    \begin{tabularx}{\columnwidth}{|m{0.40\linewidth}|m{0.10\linewidth}|X|X|}
    \hline
        Model & Rank & Rank (EN) & Rank (South Slavic) \\ \hline
        GPT-5 & 2.29 & 1.33 & 2.55 \\ \hline
        GPT-4o & 2.36 & 2.00 & 2.45 \\ \hline
        Fine-Tuned BERT-Like Model & 3.21 & 4.67 & 2.82 \\ \hline
        Gemini 2.5 Flash & 3.50 & 3.33 & 3.55 \\ \hline
        Mistral Medium 3.1 & 5.36 & 5.00 & 5.45 \\ \hline
        Gemma 3 & 5.71 & 5.67 & 5.73 \\ \hline
        LLaMA 3.3 & 6.00 & 6.67 & 5.82 \\ \hline
        Qwen 3 & 7.43 & 7.00 & 7.55 \\ \hline
        GPT-3.5-Turbo & 8.79 & 9.00 & 8.73 \\ \hline
        DeepSeek-R1-Distill & 10.00 & 10.00 & 10.00 \\ \hline
    \end{tabularx}
    \caption{Comparison of models based on their average rank (1 = best-performing, 10 = worst-performing) across all test datasets (first column), and averaged across English (second column) or South Slavic (third column) test datasets.     \label{tab:ranking}
}
\end{table}

\begin{comment}
% Old results, before updated experiments with new genre prompt
\begin{table}[!ht]
    \centering
    \begin{tabularx}{\columnwidth}{|m{0.40\linewidth}|m{0.10\linewidth}|X|X|}
    \hline
        Model & Rank & Rank (EN) & Rank (South Slavic) \\ \hline
        GPT-5 & 2.14 & 1.33 & 2.36 \\ \hline
        GPT-4o & 2.79 & 2.33 & 2.91 \\ \hline
        Fine-Tuned BERT-Like Model & 3.14 & 4.33 & 2.82 \\ \hline
        Gemini 2.5 Flash & 3.57 & 3.67 & 3.55 \\ \hline
        Mistral Medium 3.1 & 5.29 & 5.00 & 5.36 \\ \hline
        Gemma 3 & 5.64 & 6.00 & 5.55 \\ \hline
        LLaMA 3.3 & 6.29 & 6.33 & 6.27 \\ \hline
        Qwen 3 & 7.43 & 7.33 & 7.45 \\ \hline
        GPT-3.5-Turbo & 8.64 & 8.67 & 8.64 \\ \hline
        DeepSeek-R1-Distill & 10.00 & 10.00 & 10.00 \\ \hline
    \end{tabularx}
    \caption{Comparison of models based on their average rank (1 = best-performing, 10 = worst-performing) across all test datasets (first column), and averaged across English (second column) or South Slavic (third column) test datasets.     \label{tab:ranking}
}
\end{table}
\end{comment}

\begin{figure*}[htbp]
    \centering

    \begin{subfigure}[b]{0.45\textwidth}
        \centering
        \includegraphics[width=\textwidth]{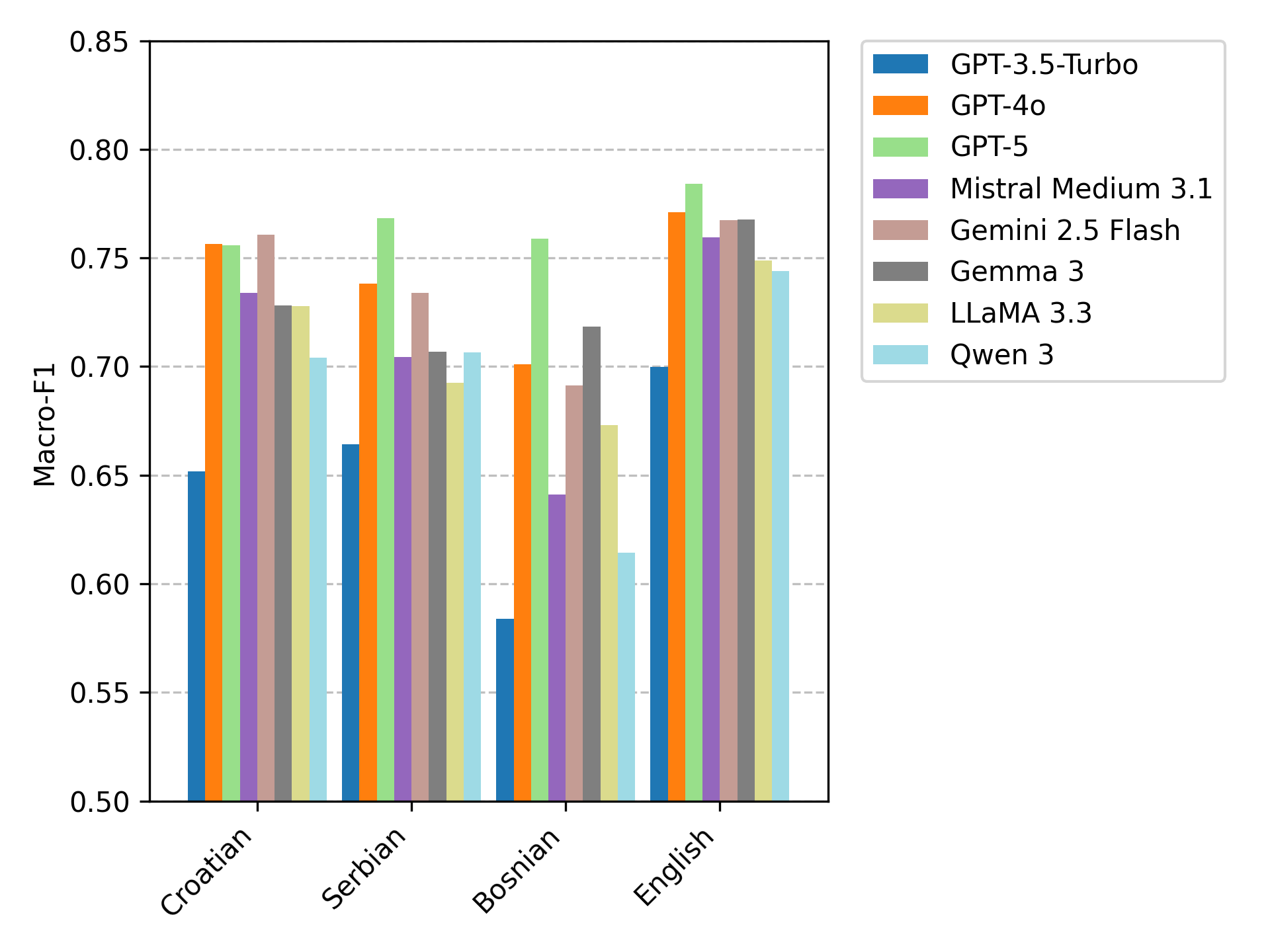}
        \caption{Sentiment classification.}
        \label{fig:sent-gpt-comparison}
    \end{subfigure}
    \hfill
    \begin{subfigure}[b]{0.45\textwidth}
        \centering
        \includegraphics[width=\textwidth]{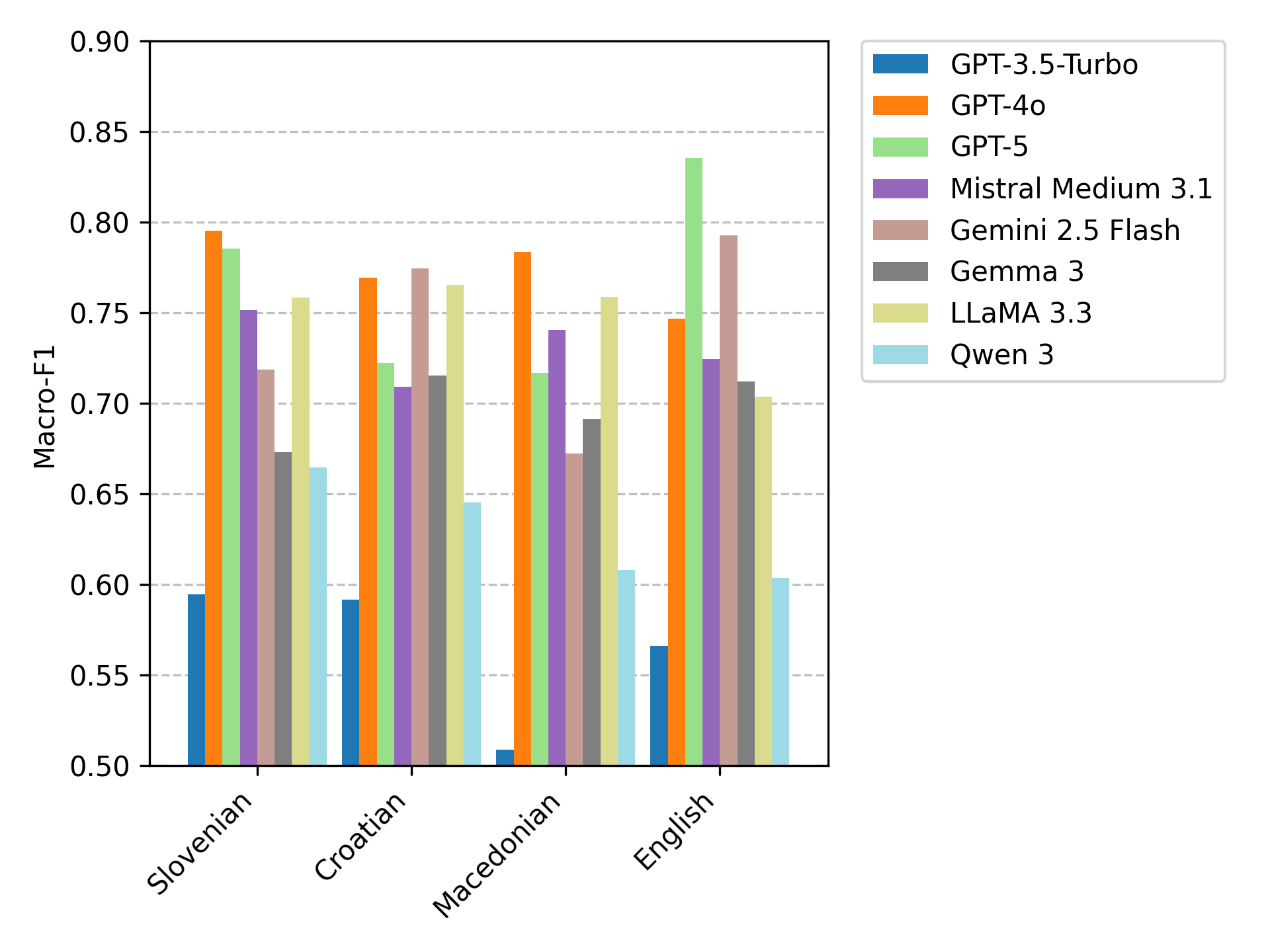}
        \caption{Automatic genre identification.}
        \label{fig:genre-gpt-comparison}
    \end{subfigure}

    \vskip\baselineskip 

    % Second row
    \begin{subfigure}[b]{0.45\textwidth}
        \centering
        \includegraphics[width=\textwidth]{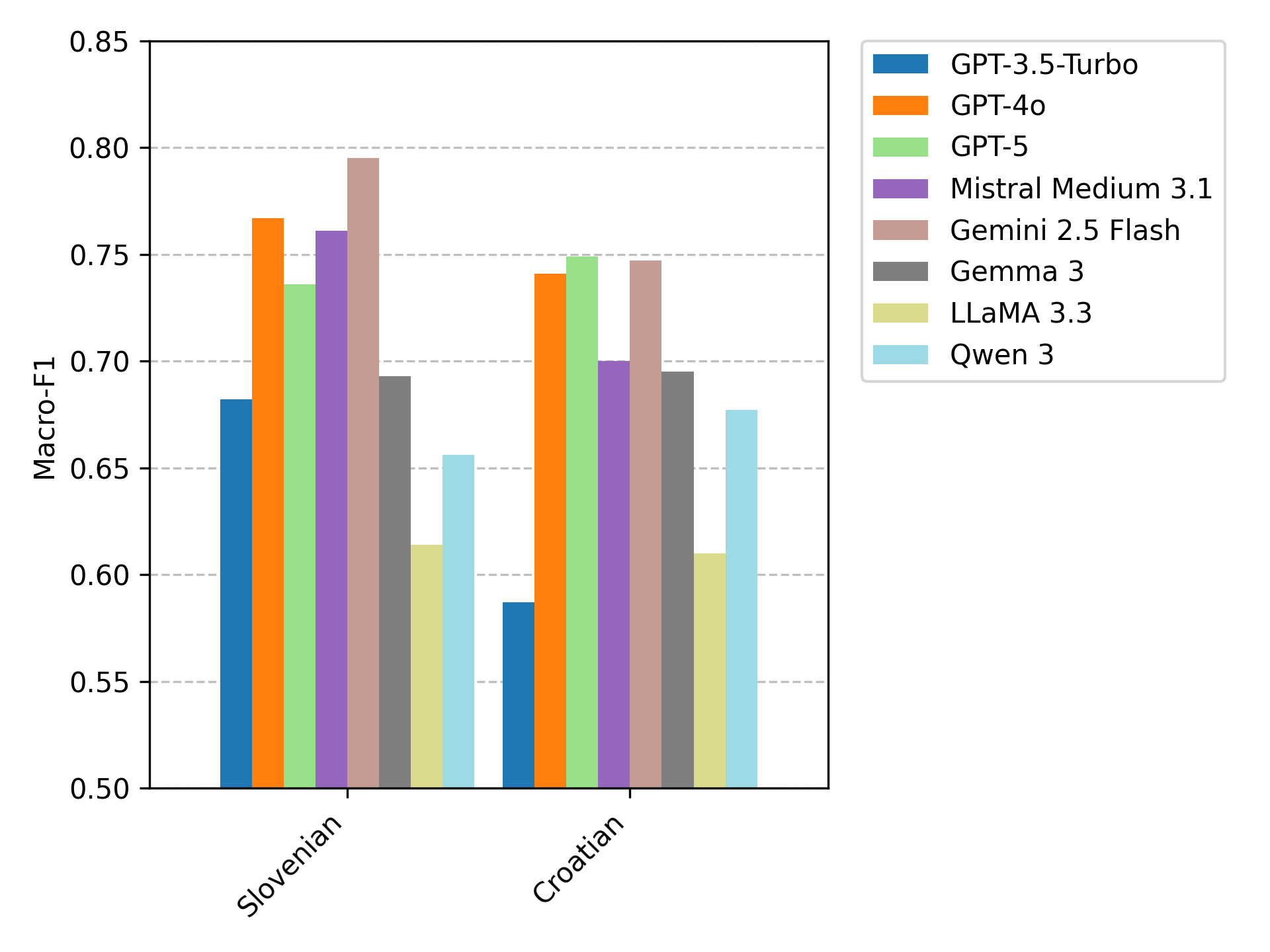}
        \caption{News topic classification.}
        \label{fig:topic-gpt-comparison}
    \end{subfigure}
    \hfill
    \begin{subfigure}[b]{0.45\textwidth}
        \centering
        \includegraphics[width=\textwidth]{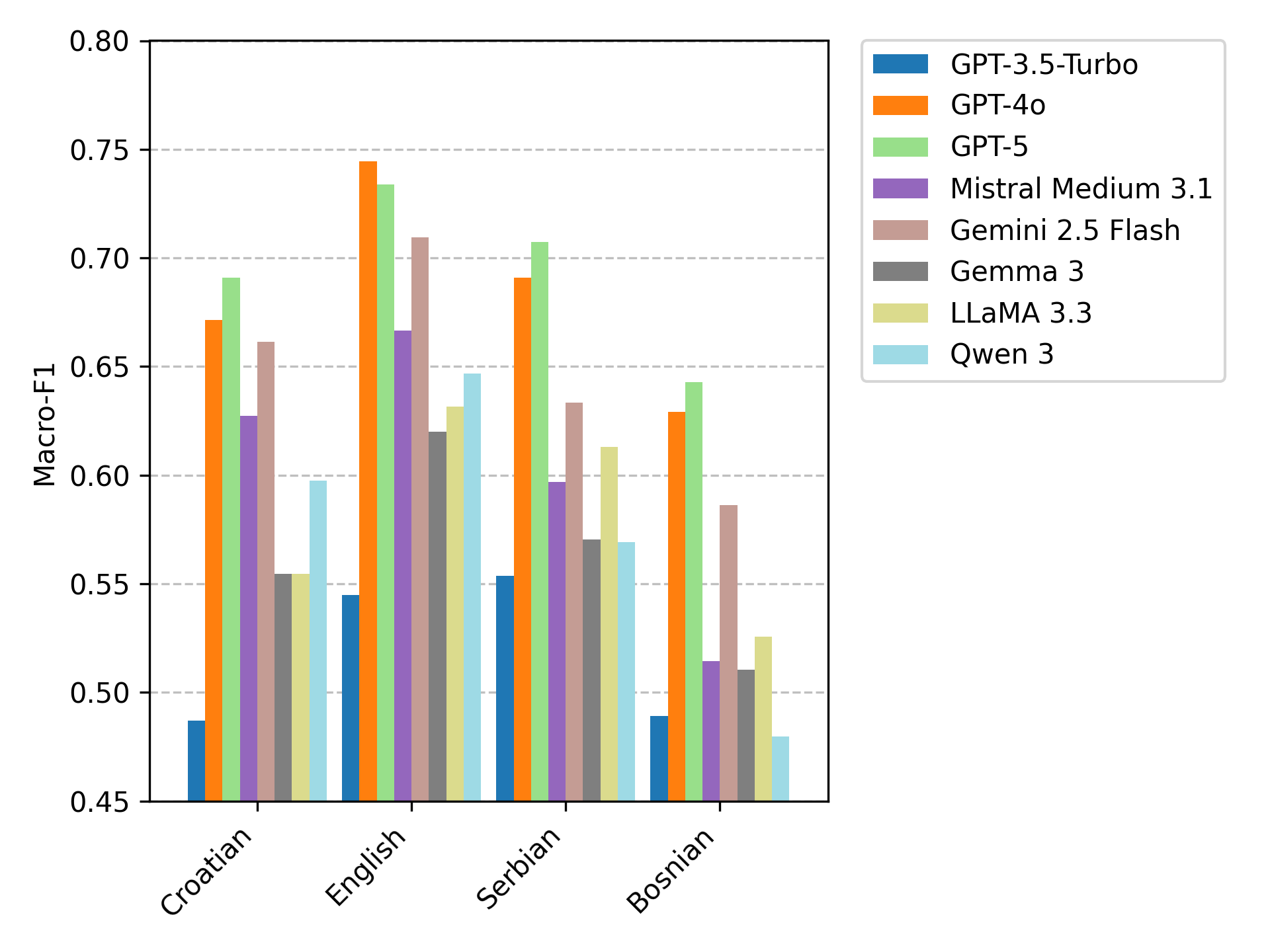}
        \caption{Parliamentary topic classification.}
        \label{fig:parlacap-gpt-comparison}
    \end{subfigure}
    
    \caption{Comparison of LLMs used in a zero-shot prompting fashion on sentiment identification (Figure \ref{fig:sent-gpt-comparison}), automatic genre identification (Figure \ref{fig:genre-gpt-comparison}), and topic classification on news (Figure \ref{fig:topic-gpt-comparison}) and parliamentary speeches (Figure \ref{fig:parlacap-gpt-comparison}).}
    \label{fig:gpt-comparison}
\end{figure*}

\subsection{State of the Art in Text Classification Tasks}
\label{sec:sota-results}

\begin{comment}
\begin{figure}[!ht]
\begin{center}
\includegraphics[width=\columnwidth]{figures/sentiment-results-heatmap.png}
\caption{Micro- and macro-F1 scores across models and languages on the sentiment classification test datasets.}
\label{fig:sentiment-results}
\end{center}
\end{figure}

\begin{figure}[!ht]
\begin{center}
\includegraphics[width=\columnwidth]{figures/genre-results-heatmap.png}
\caption{Micro- and macro-F1 scores across models and languages on the automatic genre identification test datasets.} %Instances annotated as "Other" are removed from the EN-GINCO dataset.}
\label{fig:genre-results}
\end{center}
\end{figure}

\begin{figure}[!ht]
\begin{center}
\includegraphics[width=\columnwidth]{figures/topic-classification-results.png}
\caption{Micro- and macro-F1 scores across models and languages on the news topic test datasets.}
\label{fig:iptc-results}
\end{center}
\end{figure}

\begin{figure}[!ht]
\begin{center}
\includegraphics[width=\columnwidth]{figures/parlacap-topic-results-heatmap.png}
\caption{Micro- and macro-F1 scores across models and languages on the parliamentary topic test datasets.}
\label{fig:parlacap-results}
\end{center}
\end{figure}

\end{comment}

Figure \ref{fig:main-results} provides results of model evaluation on our selection of text classification tasks. A consistent pattern emerges across all four benchmark families: LLMs, when used in a zero-shot prompting setup, achieve some of the highest scores. As shown in Table~\ref{tab:ranking}, which compares model rankings across tasks, LLMs achieve first place more often on average than the fine-tuned BERT-like model.
%Figure \ref{fig:main-results} provides more insights into the performance of BERT-like models and LLMs across the three text classification tasks. In genre and topic classification, fine-tuned BERT-like models also rank among the top performers. In contrast, in the sentiment classification task, both closed-source and open-source LLMs outperform the fine-tuned BERT-like model. The best models achieve micro- and macro-F1 scores up to approximately 0.80 on all tasks, except on topic classification on parliamentary speeches where the highest scores are between 0.65 and 0.75.

Figure \ref{fig:sentiment-results} shows that both open-weight and closed-source LLMs, used in a zero-shot prompting setup on the sentiment identification task, achieve performance that is comparable or even significantly higher to that of a fine-tuned BERT-like model trained on a large manually-annotated sentiment dataset. The only models that consistently perform worse than the fine-tuned BERT-like model are GPT-3.5-Turbo and DeepSeek-R1-Distill. Sentiment classification appears broad enough that more potent LLMs can interpret label definitions effectively without task-specific fine-tuning, reducing the benefit of additional training. 

In contrast, fine-tuned BERT-like models outperform most LLMs on automatic genre identification and topic classification tasks. These tasks depend on predefined label sets based on specific guidelines, and the strong performance of fine-tuned BERT-like models indicates that domain-specific fine-tuning on labelled data still offers an advantage over the general knowledge leveraged by LLMs in zero-shot setups. This advantage is particularly clear in genre identification for South Slavic texts, where the fine-tuned BERT-like model significantly outperforms LLMs. The likely reason for the fine-tuned model's very strong performance on South Slavic genre datasets is the curated nature of the test data -- more challenging examples were removed before and during manual annotation, unlike in the English genre test dataset where the instances were randomly sampled from an English web corpus. Nevertheless, despite this limitation, the South Slavic test dataset remains valuable for comparing the performance of LLMs.

To conclude, since some LLMs used in a zero-shot prompting setup achieve higher or comparable results to fine-tuned BERT-like models across all classification tasks and languages, as shown in Table \ref{tab:ranking}, we can confirm hypothesis H1, which proposed that zero-shot prompting with LLMs can perform comparably to fine-tuned BERT-like models.

\begin{comment}

\begin{figure}[!ht]
\begin{center}
\includegraphics[width=\columnwidth]{figures/sentiment_gpt_comparison.png}
\caption{Comparison of instruction-tuned GPT models used in a zero-shot prompting fashion on the task of sentiment identification.}
\label{fig:sent-gpt-comparison}
\end{center}
\end{figure}

\begin{figure}[!ht]
\begin{center}
\includegraphics[width=\columnwidth]{figures/genre_gpt_comparison.png}
\caption{Comparison of instruction-tuned GPT models used in a zero-shot prompting fashion on the task of automatic genre identification.}
\label{fig:genre-gpt-comparison}
\end{center}
\end{figure}

\begin{figure}[!ht]
\begin{center}
\includegraphics[width=\columnwidth]{figures/topic_gpt_comparison.png}
\caption{Comparison of instruction-tuned GPT models used in a zero-shot prompting fashion on the task of news topic classification.}
\label{fig:topic-gpt-comparison}
\end{center}
\end{figure}

\begin{figure}[!ht]
\begin{center}
\includegraphics[width=\columnwidth]{figures/parlacap_gpt_comparison.png}
\caption{Comparison of instruction-tuned GPT models used in a zero-shot prompting fashion on the task of parliamentary topic classification.}
\label{fig:parlacap-gpt-comparison}
\end{center}
\end{figure}

\end{comment}

\subsection{Comparison of Large Language Models}
\label{sec:gpt-comparison}

Figure \ref{fig:gpt-comparison} shows the performance of open-weight and closed-source LLMs, used via prompting, on the tasks of sentiment identification, automatic genre identification, news topic classification, and parliamentary topic classification. The DeepSeek-R1-Distill model is not included in the comparison, as it performs significantly worse than other models, as shown in Figure \ref{fig:main-results}.

While different models perform best across different languages and test datasets, a clear trend emerges: the top-performing models across all four benchmark families are the closed-source GPT-4o and GPT-5 from OpenAI, along with Gemini 2.5 Flash. Although GPT-5 is newer and reportedly more powerful, it does not outperform GPT-4o on all benchmarks. Among open-weight models, Gemma 3 generally achieves the best results in sentiment identification (Figure~\ref{fig:sent-gpt-comparison}) and news topic classification (Figure~\ref{fig:topic-gpt-comparison}). For automatic genre identification (Figure~\ref{fig:genre-gpt-comparison}) and parliamentary topic classification (Figure~\ref{fig:parlacap-gpt-comparison}), rankings of open-weight models vary by language. Overall, the weakest performance is observed with the older closed-source GPT-3.5-Turbo model, highlighting the rapid progress in both open-weight and closed-source model development.

\subsection{Advantages and Disadvantages of LLMs}
\label{sec:advantages-llms}

\begin{figure}[!ht]
\begin{center}
\includegraphics[width=\columnwidth]{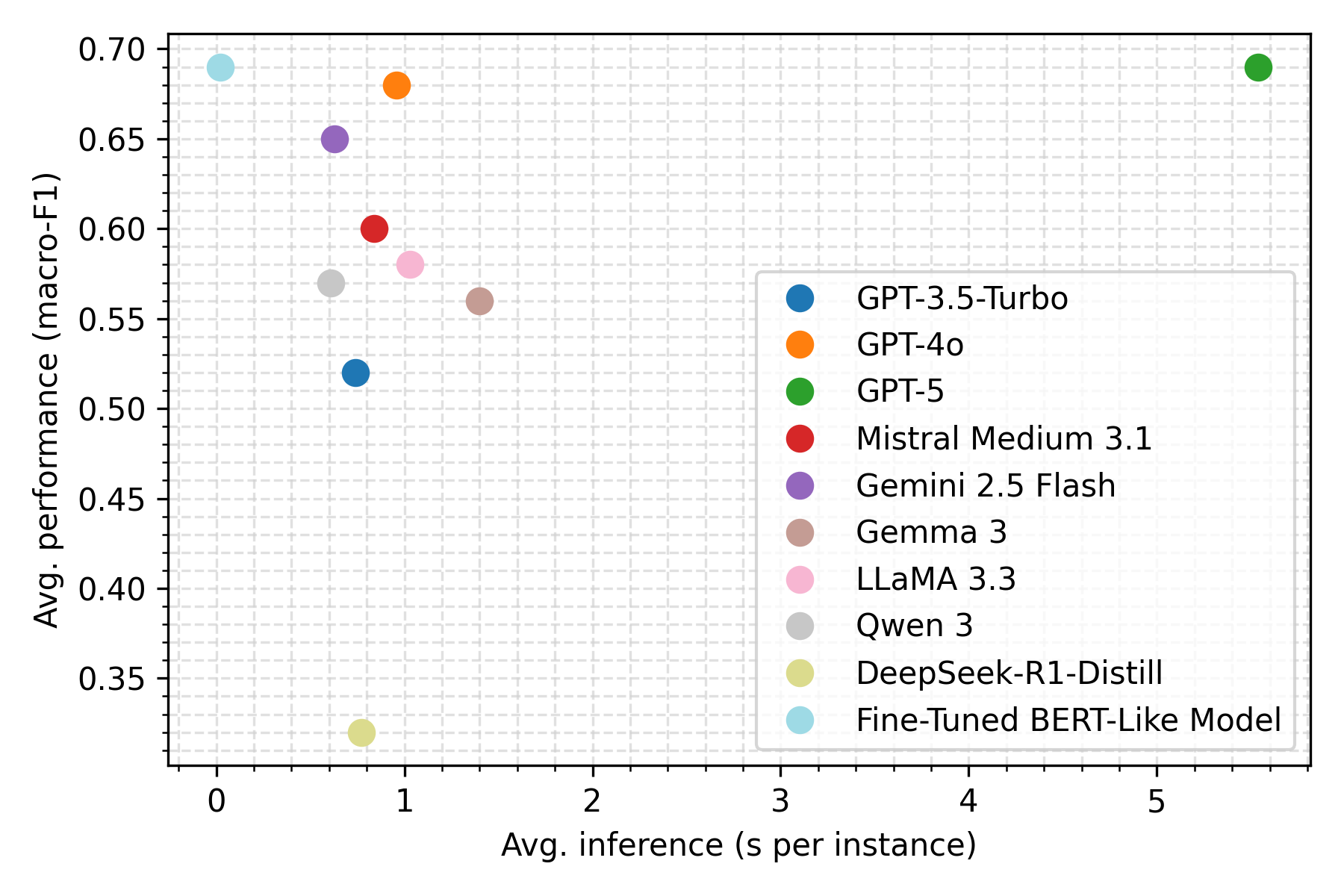}
\caption{Comparison of models on the parliamentary topic classification based on their inference speed (seconds per instance) and performance (macro-F1 scores), both averaged across all four languages.}
\label{fig:parlacap-inference}
\end{center}
\end{figure}

A clear advantage of LLMs is that they do not require manually-annotated training data for specific tasks, yet still achieve strong performance when provided only with task instructions and brief label descriptions. However, these models are significantly more computationally expensive than fine-tuned BERT-like models. While closed-source models deliver the best performance, as shown in previous sections, they come with several limitations: they are costly to use, their architectures and pre-training data are not publicly disclosed, and access through APIs hinders reproducibility, in contrast to open-weight LLMs and fine-tuned BERT-like models.

What is more, the inference speed of all LLMs is significantly slower than that of a fine-tuned BERT-like model. As shown in Figure \ref{fig:parlacap-inference}, the fine-tuned BERT-like model achieves one of the highest macro-F1 scores on the topic classification task for parliamentary speeches, while maintaining a very low inference time of just 0.02 seconds per instance. In contrast, most LLMs have inference times between 0.6 and 1.4 seconds per instance, making them three to seven times slower for annotating the same dataset. The slowest model, GPT-5, takes 5.5 seconds per instance, which renders it impractical for large-scale automatic annotation of text collections. In this regard, fine-tuned BERT-like models offer a key advantage due to their lower computational cost and higher inference speed. Moreover, they can be trained on training data that is annotated by LLMs using the recently introduced LLM teacher-student paradigm \citep{kuzman-iptc-classification}, which considerably reduces the effort needed to develop task-specific models.

Another limitation of LLMs, as revealed by the experiments, is their occasional deviation from the defined label set. This issue was especially noticeable in topic classification and, to a lesser extent, in genre identification. The highest rate of label hallucination was found in the DeepSeek-R1-Distill model, which produced non-existing labels for 8\% of instances in the news topic test datasets and 4\% in the genre test dataset. Similar issues were also observed, though much less frequently (less than 1\%), with the LLaMA 3.3, Gemma 3, Qwen 3 and Mistral Medium 3.1 models. In contrast, fine-tuned BERT-like models do not suffer from this issue, as they output probabilities for the predefined classes.

\begin{table}[!ht]
    \centering
    \begin{tabularx}{\columnwidth}{|X|X|X|}
    \hline
        Model & Difference (sentiment) & Difference (topic) \\ \hline
        GPT-5 & 0.02 & 0.05 \\ \hline
        GPT-4o & 0.04 & 0.08 \\ \hline
        Gemini 2.5 Flash & 0.04 & 0.08 \\ \hline
        Gemma 3 & 0.05 & 0.07 \\ \hline
        LLaMA 3.3 & 0.05 & 0.07 \\ \hline
        Mistral Medium 3.1 & 0.07 & 0.09 \\ \hline
        Qwen 3 & 0.07 & 0.10 \\ \hline
        GPT-3.5-Turbo & 0.07 & 0.03 \\ \hline
    \end{tabularx}
    \caption{Difference between model performance in macro-F1 scores obtained on sentiment and topic classification in parliamentary texts on English versus the average macro-F1 scores on South Slavic languages.}
    \label{tab:sent-en-vs-south-slavic}
\end{table}

\subsection{Performance on English versus on South Slavic languages}
\label{sec:results-en-vs-south-slavic}

The sentiment identification ParlaSent and the topic classification ParlaCAP benchmark families comprise test datasets in South Slavic languages and English that were constructed with the same methodology. Thus, they also allow for a comparison of the performance of the LLMs on English, a highly resourced language, with South Slavic languages, which are significantly less represented in the pretraining and instruction-tuning datasets used to develop large language models.

As shown in Table \ref{tab:sent-en-vs-south-slavic}, the differences in macro-F1 scores between English and the average of macro-F1 scores for South Slavic languages are relatively small for sentiment identification, ranging from 2 to 7 points. For topic classification, the performance gap is slightly larger, ranging from 3 to 10 points. This is likely due to the increased difficulty of the task, which involves greater label granularity: 22 labels compared to just 3 in sentiment classification. These findings partially confirm hypothesis H2, which stated that LLMs, when used in a zero-shot setup, perform comparably on text classification tasks in South Slavic languages as they do on English.

Interestingly, even the open-weight LLaMA 3.3 model -- reported to support only eight languages, excluding the South Slavic group -- does not show a substantial performance drop when applied to South Slavic languages compared to English.

\section{Conclusion}

In this paper, we evaluated how well current machine learning technologies handle text classification tasks in South Slavic languages. We compared fine-tuned BERT-like models with decoder-only generative large language models (LLMs) that are used in a zero-shot prompting setup across three tasks and three text domains: sentiment classification in parliamentary texts, news topic classification, topic classification in parliamentary texts, and automatic genre identification on web texts.

Our results show that LLMs used with prompting, where only a brief task description and labels were provided, achieved strong results across all tasks and languages, particularly the closed-source GPT-4o \citep{openai-gpt4o}, GPT-5 \citep{gpt-5} and Gemini 2.5 Flash \citep{comanici2025gemini} models. The performance of LLMs is comparable or higher to that of fine-tuned BERT-like models specialized for the tasks. On the sentiment identification task, most open-weight and closed-source LLMs outperformed the fine-tuned model, demonstrating strong general knowledge of the notion of sentiment. For genre and topic classification, however, fine-tuning BERT-like models remains beneficial, as these tasks rely on predefined label sets and fine-tuning aligns the models more closely with the task requirements.

Interestingly, LLMs perform similarly in English and South Slavic languages, with rather minor drops in micro- and macro-F1 scores, namely a drop of 2 to 7 points in terms of macro-F1 scores on sentiment classification, and a slightly higher drop from 3 to 10 points on topic classification in parliamentary texts. This suggests that the gap in multilingual performance is smaller than expected, even for open-weight models not explicitly dedicated to these languages. %However, performance still does not exceed 0.80 in micro- and macro-F1 scores across tasks, showing that these tasks are far from solved.

Although large language models offer impressive zero-shot performance and reduce the need for annotated data, they come with higher computational costs and are more prone to producing invalid labels. Moreover, their inference speed is at least three times slower than that of the fine-tuned BERT-like models. Thus, their use in use cases with extensive data to be processed, such as automatic enrichment of large corpora with text categories, remains impractical due to their high computational demands. In contrast, fine-tuned BERT-like models are more computationally efficient and can be better tailored to the specific characteristics of a task and its domain. They remain a practical and reliable choice for text classification tasks, especially when computational resources are limited, high inference speed is desired or output reliability is critical. Moreover, it is possible to combine the strengths of both approaches, as proposed by the LLM teacher-student paradigm \citep{kuzman-iptc-classification}: LLMs can be used to automatically annotate training data, reducing the need for costly and time-consuming manual annotation, while fine-tuned BERT-like models can then be trained on these datasets.
 
This study represents only an initial step to systematically benchmark text classification performance in South Slavic languages. Although our evaluation includes four diverse benchmark families, some of the test datasets remain relatively small. Future work will aim to increase dataset sizes, include more South Slavic languages and dialects, and introduce additional classification tasks. As new large language models continue to emerge rapidly, it will also be important to establish ongoing evaluations to track whether their performance continues to improve, particularly on South Slavic languages. Importantly, this study only evaluated the performance of LLMs in a zero-shot prompting setup. In future work, we plan to extend the evaluation to include few-shot prompting and fine-tuning on training data. To support further research and facilitate reproducibility, we have made all code, evaluation scripts, and results publicly available.\footnote{\url{https://github.com/TajaKuzman/Benchmarking-Text-Classification-on-South-Slavic}} Additionally, we have developed an interactive dashboard\footnote{See the CLASSLA LLM Evaluation Dashboard for South Slavic Languages at \url{https://www.clarin.si/classla-llm-dashboard/}.} that enables users to explore the results of our evaluation of large language models on the tasks presented in this paper, as well as on additional commonsense reasoning tasks. The dashboard is an ongoing project that monitors the performance of newly released large language models on South Slavic languages and dialects. In future work, we plan to expand both the range of tasks and the coverage of South Slavic languages and dialects included in the dashboard.

\section{Ethical Considerations and Limitations}

Our study has several limitations that should be acknowledged. First, while we aimed to include a broad set of South Slavic languages, some -- most notably Bulgarian -- were not covered in our experiments. We assume that the performance on Bulgarian would be similar to that observed for Macedonian, given their close linguistic proximity, or the results for Bulgarian could be slightly better, as Macedonian is comparatively more low-resourced. Moreover, due to the high computational cost of evaluating the LLMs on all the test datasets and the financial cost associated with the use of closed-source models, each model was evaluated on each dataset only once. This setup prevents us from fully estimating the variance of the results, however, based on our preliminary experiments, we expect this variance to be relatively small. Finally, the scope of our evaluation remains limited in terms of test datasets, language coverage and tasks. Expanding the range of benchmarks would allow for a more comprehensive validation of our findings, particularly regarding the hypothesis that LLMs can perform on par with fine-tuned BERT-like models across diverse natural language understanding tasks, languages and language varieties.

\section{Acknowledgements}

We would like to thank the developers of the llm.ijs.si service \citep{maric_local_llms} for establishing the LLM inference platform deployed at the Jožef Stefan Institute, which provided convenient access to the open-weight large language models used in this study. We also thank the annotators of the test datasets for their diligence and the time devoted to manual annotation, which resulted in the high-quality evaluation datasets used in this work. Lastly, we would like to thank the \href{https://www.clarin.si/info/k-centre/}{CLASSLA knowledge centre for South Slavic languages} and the Slovenian \href{https://www.clarin.si/info/about/}{CLARIN.SI infrastructure} for their valuable support.

% add here also ParlaCAP
This work was supported in part by the projects ``Spoken Language Resources and Speech Technologies for the Slovenian Language'' (Grant J7-4642), ``Large Language Models for Digital Humanities'' (Grant GC-0002), the research programme ``Language Resources and Technologies for Slovene'' (Grant P6-0411), all funded by the ARIS Slovenian Research and Innovation Agency, and the research project ``Embeddings-based techniques for Media Monitoring Applications'' (L2-50070), co-funded by the Kliping d.o.o. agency. The authors acknowledge the OSCARS project -- and its ParlaCAP cascading grant project --, which has received funding from the European Commission’s Horizon Europe Research and Innovation programme under grant agreement No. 101129751. %This research was supported by LLMs4EU, co-funded by the Digital Europe Programme under GA 101198470. This research is co-funded by the European Union. Views and opinions expressed are however those of the author(s) only and do not necessarily reflect those of the European Union or the European Commission. Neither the European Union nor the granting authority can be held responsible for them.

\section{Bibliographical References}
\urlstyle{same}
\bibliographystyle{lrec2026-natbib}
\bibliography{references}

\begin{thebibliography}{55}
\expandafter\ifx\csname natexlab\endcsname\relax\def\natexlab#1{#1}\fi

\bibitem[{Ba{\~n}{\'o}n et~al.(2022)Ba{\~n}{\'o}n, Espl{\`a}-Gomis, Forcada, Garc{\'\i}a-Romero, Kuzman, Ljube{\v{s}}i{\'c}, van Noord, Sempere, Ram{\'\i}rez-S{\'a}nchez, Rupnik et~al.}]{banon2022macocu}
Marta Ba{\~n}{\'o}n, Miquel Espl{\`a}-Gomis, Mikel~L Forcada, Cristian Garc{\'\i}a-Romero, Taja Kuzman, Nikola Ljube{\v{s}}i{\'c}, Rik van Noord, Leopoldo~Pla Sempere, Gema Ram{\'\i}rez-S{\'a}nchez, Peter Rupnik, et~al. 2022.
\newblock \href {https://aclanthology.org/2022.eamt-1.41/} {{MaCoCu: Massive collection and curation of monolingual and bilingual data: focus on under-resourced languages}}.
\newblock In \emph{23rd Annual Conference of the European Association for Machine Translation}, pages 301--302.

\bibitem[{Baumgartner et~al.(2019)Baumgartner, Breunig, and Grossman}]{baumgartner2019comparative}
Frank~R Baumgartner, Christian Breunig, and Emiliano Grossman. 2019.
\newblock \emph{{Comparative Policy Agendas: Theory, Tools, Data}}.
\newblock Oxford University Press.

\bibitem[{Bevan(2019)}]{bevan2019gone}
Shaun Bevan. 2019.
\newblock {Gone Fishing}.
\newblock \emph{Comparative Policy Agendas: Theory, Tools, Data}, pages 17--34.

\bibitem[{Comanici et~al.(2025)Comanici, Bieber, Schaekermann, Pasupat, Sachdeva, Dhillon, Blistein, Ram, Zhang, Rosen et~al.}]{comanici2025gemini}
Gheorghe Comanici, Eric Bieber, Mike Schaekermann, Ice Pasupat, Noveen Sachdeva, Inderjit Dhillon, Marcel Blistein, Ori Ram, Dan Zhang, Evan Rosen, et~al. 2025.
\newblock \href {https://arxiv.org/abs/2507.06261} {Gemini 2.5: Pushing the frontier with advanced reasoning, multimodality, long context, and next generation agentic capabilities}.
\newblock \emph{arXiv preprint arXiv:2507.06261}.

\bibitem[{Conneau et~al.(2020)Conneau, Khandelwal, Goyal, Chaudhary, Wenzek, Guzm{\'a}n, Grave, Ott, Zettlemoyer, and Stoyanov}]{conneau2020unsupervised}
Alexis Conneau, Kartikay Khandelwal, Naman Goyal, Vishrav Chaudhary, Guillaume Wenzek, Francisco Guzm{\'a}n, {\'E}douard Grave, Myle Ott, Luke Zettlemoyer, and Veselin Stoyanov. 2020.
\newblock \href {https://aclanthology.org/P19-4007.pdf} {{Unsupervised Cross-lingual Representation Learning at Scale}}.
\newblock In \emph{{58th Annual Meeting of the Association for Computational Linguistics}}, pages 8440--8451.

\bibitem[{De~Clercq et~al.(2020)De~Clercq, De~Bruyne, and Hoste}]{de2020news}
Orph{\'e}e De~Clercq, Luna De~Bruyne, and V{\'e}ronique Hoste. 2020.
\newblock \href {https://clinjournal.org/clinj/article/view/103} {News topic classification as a first step towards diverse news recommendation}.
\newblock \emph{Computational Linguistics in the Netherlands Journal}, 10:37--55.

\bibitem[{Erjavec et~al.(2025)Erjavec, Kopp, Ljube{\v{s}}i{\'c}, Kuzman, Rayson, Osenova, Ogrodniczuk, {\c{C}}{\"o}ltekin, Kor{\v{z}}inek, Meden et~al.}]{erjavec2025parlamint}
Toma{\v{z}} Erjavec, Maty{\'a}{\v{s}} Kopp, Nikola Ljube{\v{s}}i{\'c}, Taja Kuzman, Paul Rayson, Petya Osenova, Maciej Ogrodniczuk, {\c{C}}a{\u{g}}r{\i} {\c{C}}{\"o}ltekin, Danijel Kor{\v{z}}inek, Katja Meden, et~al. 2025.
\newblock \href {https://link.springer.com/article/10.1007/s10579-024-09798-w} {{ParlaMint II: advancing comparable parliamentary corpora across Europe}}.
\newblock \emph{Language Resources and Evaluation}, 59(3):2071--2102.

\bibitem[{Erjavec et~al.(2024)Erjavec, Kopp, Ogrodniczuk, Osenova et~al.}]{parlamint_41}
Toma{\v z} Erjavec, Maty{\'a}{\v s} Kopp, Maciej Ogrodniczuk, Petya Osenova, et~al. 2024.
\newblock \href {http://hdl.handle.net/11356/1912} {Multilingual comparable corpora of parliamentary debates {ParlaMint} 4.1}.
\newblock Slovenian language resource repository {CLARIN}.{SI}: http://hdl.handle.net/11356/1912.

\bibitem[{{Gemma Team} et~al.(2025){Gemma Team}, Kamath, Ferret, Pathak, Vieillard, Merhej, Perrin, Matejovicova, Ram{\'e}, Rivi{\`e}re et~al.}]{gemma3}
{Gemma Team}, Aishwarya Kamath, Johan Ferret, Shreya Pathak, Nino Vieillard, Ramona Merhej, Sarah Perrin, Tatiana Matejovicova, Alexandre Ram{\'e}, Morgane Rivi{\`e}re, et~al. 2025.
\newblock \href {https://arxiv.org/pdf/2503.19786} {Gemma 3 technical report}.
\newblock \emph{arXiv preprint arXiv:2503.19786}.

\bibitem[{Guo et~al.(2025)Guo, Yang, Zhang, Song, Zhang, Xu, Zhu, Ma, Wang, Bi et~al.}]{guo2025deepseek}
Daya Guo, Dejian Yang, Haowei Zhang, Junxiao Song, Ruoyu Zhang, Runxin Xu, Qihao Zhu, Shirong Ma, Peiyi Wang, Xiao Bi, et~al. 2025.
\newblock \href {https://arxiv.org/pdf/2501.12948?} {{DeepSeek-R1: Incentivizing Reasoning Capability in LLMs via Reinforcement Learning}}.
\newblock \emph{arXiv preprint arXiv:2501.12948}.

\bibitem[{Hendy et~al.(2023)Hendy, Abdelrehim, Sharaf, Raunak, Gabr, Matsushita, Kim, Afify, and Awadalla}]{hendy2023good}
Amr Hendy, Mohamed Abdelrehim, Amr Sharaf, Vikas Raunak, Mohamed Gabr, Hitokazu Matsushita, Young~Jin Kim, Mohamed Afify, and Hany~Hassan Awadalla. 2023.
\newblock \href {https://arxiv.org/abs/2302.09210} {{How Good Are GPT Models at Machine Translation? A Comprehensive Evaluation}}.
\newblock \emph{arXiv preprint arXiv:2302.09210}.

\bibitem[{Huang et~al.(2023)Huang, Kwak, and An}]{huang2023chatgpt}
Fan Huang, Haewoon Kwak, and Jisun An. 2023.
\newblock \href {https://dl.acm.org/doi/abs/10.1145/3543873.3587368} {{Is ChatGPT better than human annotators? Potential and limitations of ChatGPT in explaining implicit hate speech}}.
\newblock In \emph{{Companion Proceedings of the ACM Web Conference 2023}}, pages 294--297.

\bibitem[{IPTC(2022)}]{iptcGroupsNewsCodes}
IPTC. 2022.
\newblock \href {https://iptc.org/standards/newscodes/groups/#descrncd} {{G}roups of {N}ews{C}odes}.
\newblock \url{https://iptc.org/standards/newscodes/groups/#descrncd}.
\newblock Accessed October 29, 2024.

\bibitem[{Jakub{\'\i}{\v{c}}ek et~al.(2013)Jakub{\'\i}{\v{c}}ek, Kilgarriff, Kov{\'a}{\v{r}}, Rychl{\`y}, and Suchomel}]{jakubivcek2013tenten}
Milo{\v{s}} Jakub{\'\i}{\v{c}}ek, Adam Kilgarriff, Vojt{\v{e}}ch Kov{\'a}{\v{r}}, Pavel Rychl{\`y}, and V{\'\i}t Suchomel. 2013.
\newblock \href {http://www.sketchengine.co.uk/wp-content/uploads/The_TenTen_Corpus_2013.pdf} {{The TenTen corpus family}}.
\newblock In \emph{7th international corpus linguistics conference CL}, pages 125--127. Lancaster University.

\bibitem[{Kenton and Toutanova(2019)}]{bert}
Jacob Devlin Ming-Wei~Chang Kenton and Lee~Kristina Toutanova. 2019.
\newblock \href {https://aclanthology.org/N19-1423.pdf} {{BERT: Pre-training of Deep Bidirectional Transformers for Language Understanding}}.
\newblock \emph{Proceedings of NAACL-HLT}, pages 4171--4186.

\bibitem[{Kostina et~al.(2025)Kostina, Dikaiakos, Stefanidis, and Pallis}]{kostina2025large}
Arina Kostina, Marios~D Dikaiakos, Dimosthenis Stefanidis, and George Pallis. 2025.
\newblock \href {https://arxiv.org/abs/2501.08457} {{Large language models for text classification: Case study and comprehensive review}}.
\newblock \emph{arXiv preprint arXiv:2501.08457}.

\bibitem[{Krippendorff(2018)}]{krippendorff2018content}
Klaus Krippendorff. 2018.
\newblock \emph{{Content analysis: An introduction to its methodology}}.
\newblock Sage Publications.

\bibitem[{Kuzman and Ljube{\v{s}}i{\'c}(2023)}]{kuzman2023survey}
Taja Kuzman and Nikola Ljube{\v{s}}i{\'c}. 2023.
\newblock \href {https://link.springer.com/content/pdf/10.1007/s10579-023-09695-8.pdf} {Automatic genre identification: a survey}.
\newblock \emph{Language Resources and Evaluation}, pages 1--34.

\bibitem[{Kuzman and Ljube{\v s}i{\'c}(2023)}]{x-genre-classifier-huggingface}
Taja Kuzman and Nikola Ljube{\v s}i{\'c}. 2023.
\newblock \href {https://doi.org/10.57967/hf/0927} {{Multilingual text genre classification model X-GENRE}}.
\newblock {Hugging Face}: www.doi.org/10.57967/hf/0927.

\bibitem[{Kuzman and Ljube{\v s}i{\'c}(2024{\natexlab{a}})}]{x-genre-dataset}
Taja Kuzman and Nikola Ljube{\v s}i{\'c}. 2024{\natexlab{a}}.
\newblock \href {http://hdl.handle.net/11356/1960} {{English-Slovenian text genre dataset X-{GENRE}}}.
\newblock {Slovenian Language Resource Repository {CLARIN}.{SI}}: http://hdl.handle.net/11356/1960.

\bibitem[{Kuzman and Ljube{\v s}i{\'c}(2024{\natexlab{b}})}]{macocu-genre}
Taja Kuzman and Nikola Ljube{\v s}i{\'c}. 2024{\natexlab{b}}.
\newblock \href {http://hdl.handle.net/11356/1969} {{Genre-enriched web corpora {MaCoCu}-Genre}}.
\newblock {Slovenian Language Resource Repository {CLARIN}.{SI}}.

\bibitem[{Kuzman and Ljube{\v s}i{\'c}(2024{\natexlab{c}})}]{emmediatopic}
Taja Kuzman and Nikola Ljube{\v s}i{\'c}. 2024{\natexlab{c}}.
\newblock \href {http://hdl.handle.net/11356/1991} {{Multilingual IPTC Media Topic dataset EMMediaTopic 1.0}}.
\newblock {Slovenian Language Resource Repository {CLARIN}.{SI}}: http://hdl.handle.net/11356/1991.

\bibitem[{Kuzman and Ljube{\v s}i{\'c}(2024{\natexlab{d}})}]{x-genre-repository}
Taja Kuzman and Nikola Ljube{\v s}i{\'c}. 2024{\natexlab{d}}.
\newblock \href {http://hdl.handle.net/11356/1961} {{Multilingual text genre classification model X-{GENRE}}}.
\newblock Slovenian language resource repository {CLARIN}.{SI}: http://hdl.handle.net/11356/1961.

\bibitem[{Kuzman and Ljube{\v s}i{\'c}(2025)}]{iptc_model_huggingface}
Taja Kuzman and Nikola Ljube{\v s}i{\'c}. 2025.
\newblock \href {https://huggingface.co/classla/multilingual-IPTC-news-topic-classifier} {{ Multilingual IPTC News Topic Classifier}}.
\newblock {Hugging Face}: https://doi.org/10.57967/hf/4709.

\bibitem[{Kuzman and Ljubešić(2025)}]{kuzman-iptc-classification}
Taja Kuzman and Nikola Ljubešić. 2025.
\newblock \href {https://doi.org/10.1109/ACCESS.2025.3544814} {{LLM Teacher-Student Framework for Text Classification With No Manually Annotated Data: A Case Study in IPTC News Topic Classification}}.
\newblock \emph{IEEE Access}, 13:35621--35633.

\bibitem[{Kuzman et~al.(2023)Kuzman, Mozeti{\v{c}}, and Ljube{\v{s}}i{\'c}}]{kuzman2023automatic}
Taja Kuzman, Igor Mozeti{\v{c}}, and Nikola Ljube{\v{s}}i{\'c}. 2023.
\newblock \href {https://www.mdpi.com/2504-4990/5/3/59} {{Automatic Genre Identification for Robust Enrichment of Massive Text Collections: Investigation of Classification Methods in the Era of Large Language Models}}.
\newblock \emph{Machine Learning and Knowledge Extraction}, 5(3):1149--1175.

\bibitem[{Kuzman et~al.(2022)Kuzman, Rupnik, and Ljube{\v{s}}i{\'c}}]{kuzman-rupnik-ljubei:2022:LREC}
Taja Kuzman, Peter Rupnik, and Nikola Ljube{\v{s}}i{\'c}. 2022.
\newblock \href {https://aclanthology.org/2022.lrec-1.170/} {{The GINCO Training Dataset for Web Genre Identification of Documents Out in the Wild}}.
\newblock In \emph{{Language Resources and Evaluation Conference}}, pages 1584--1594, Marseille, France. European Language Resources Association.

\bibitem[{Kuzman~Punger{\v s}ek and Ljube{\v s}i{\'c}(2025)}]{parlacap_model}
Taja Kuzman~Punger{\v s}ek and Nikola Ljube{\v s}i{\'c}. 2025.
\newblock \href {https://huggingface.co/classla/ParlaCAP-Topic-Classifier} {{Multilingual ParlaCAP model for CAP Topic Classification in Parliamentary Speeches}}.
\newblock {Hugging Face}: https://doi.org/10.57967/hf/6684.

\bibitem[{Kuzman~Punger{\v s}ek and Ljube{\v s}i{\'c}(2026)}]{parlacap-train}
Taja Kuzman~Punger{\v s}ek and Nikola Ljube{\v s}i{\'c}. 2026.
\newblock \href {http://hdl.handle.net/11356/2093} {{Multilingual training dataset for CAP policy topic classification ParlaCAP-train}}.
\newblock Slovenian language resource repository {CLARIN}.{SI}: http://hdl.handle.net/11356/2093.

\bibitem[{Kuzman~Punger{\v s}ek et~al.(2026)Kuzman~Punger{\v s}ek, Rupnik, {\v S}irini{\'c}, and Ljube{\v s}i{\'c}}]{pungersek2026parlacap-paper}
Taja Kuzman~Punger{\v s}ek, Peter Rupnik, Daniela {\v S}irini{\'c}, and Nikola Ljube{\v s}i{\'c}. 2026.
\newblock \href {http://arxiv.org/abs/2602.16516} {{Supercharging Agenda Setting Research: The ParlaCAP Dataset of 28 European Parliaments and a Scalable Multilingual LLM-Based Classification}}.
\newblock \emph{arXiv preprint arXiv:2602.16516}.

\bibitem[{Ljube{\v{s}}i{\'c} et~al.(2024{\natexlab{a}})Ljube{\v{s}}i{\'c}, Galant, Ben{\v{c}}ina, {\v{C}}ibej, Milosavljevi{\'c}, Rupnik, and Kuzman}]{ljubevsic2024dialect}
Nikola Ljube{\v{s}}i{\'c}, Nada Galant, Sonja Ben{\v{c}}ina, Jaka {\v{C}}ibej, Stefan Milosavljevi{\'c}, Peter Rupnik, and Taja Kuzman. 2024{\natexlab{a}}.
\newblock \href {https://aclanthology.org/2024.vardial-1.7.pdf} {{DIALECT-COPA: Extending the Standard Translations of the COPA Causal Commonsense Reasoning Dataset to South Slavic Dialects}}.
\newblock In \emph{Proceedings of the Eleventh Workshop on NLP for Similar Languages, Varieties, and Dialects (VarDial 2024)}, pages 89--98.

\bibitem[{Ljube{\v{s}}i{\'c} et~al.(2024{\natexlab{b}})Ljube{\v{s}}i{\'c}, Kuzman, Rupnik, Vuli{\'c}, Schmidt, and Glava{\v{s}}}]{ljubevsic2024jsi}
Nikola Ljube{\v{s}}i{\'c}, Taja Kuzman, Peter Rupnik, Ivan Vuli{\'c}, Fabian Schmidt, and Goran Glava{\v{s}}. 2024{\natexlab{b}}.
\newblock \href {https://aclanthology.org/2024.vardial-1.18.pdf} {{JSI and W{\"u}NLP at the DIALECT-COPA Shared Task: In-Context Learning From Just a Few Dialectal Examples Gets You Quite Far}}.
\newblock In \emph{Proceedings of the Eleventh Workshop on NLP for Similar Languages, Varieties, and Dialects (VarDial 2024)}, pages 209--219.

\bibitem[{Ljubešić et~al.(2023)Ljubešić, Rupnik, and van Noord}]{xlm-r-parla}
Nikola Ljubešić, Peter Rupnik, and Rik van Noord. 2023.
\newblock \href {https://huggingface.co/classla/xlm-r-parla} {{Multilingual parliamentary model XLM-R-parla}}.
\newblock {Hugging Face}: https://doi.org/10.57967/hf/6717.

\bibitem[{Marić et~al.(2025)Marić, Koloski, Demšar, Javoršek, and Džeroski}]{maric_local_llms}
Nikola Marić, Boshko Koloski, Damjan Demšar, Jan~Jona Javoršek, and Sašo Džeroski. 2025.
\newblock \href {https://ai4science.si/wp-content/uploads/2025/09/AI4Science_Book_of_Abstracts-4.pdf} {{Running large language models locally: design and operational insights with llm.ijs.si}}.
\newblock \emph{{International conference AI for science 2025: Ljubljana, Slovenia, 22.09.2025-26.09.2025}}, page~77.

\bibitem[{Meng et~al.(2022)Meng, Huang, Zhang, and Han}]{meng2022generating}
Yu~Meng, Jiaxin Huang, Yu~Zhang, and Jiawei Han. 2022.
\newblock \href {https://proceedings.neurips.cc/paper_files/paper/2022/hash/0346c148ba1c21c6b4780a961ea141dc-Abstract-Conference.html} {Generating training data with language models: Towards zero-shot language understanding}.
\newblock \emph{Advances in Neural Information Processing Systems}, 35:462--477.

\bibitem[{{Meta}(2024)}]{llama33modelcard}
{Meta}. 2024.
\newblock {Llama 3.3 Model Card}.
\newblock \url{https://github.com/meta-llama/llama-models/blob/main/models/llama3_3/MODEL_CARD.md}.
\newblock Accessed: June 26, 2025.

\bibitem[{Minaee et~al.(2020)Minaee, Kalchbrenner, Cambria, Nikzad, Chenaghlu, and Gao}]{minaee2020deep}
Shervin Minaee, Nal Kalchbrenner, Erik Cambria, Narjes Nikzad, Meysam Chenaghlu, and Jianfeng Gao. 2020.
\newblock \href {https://arxiv.org/pdf/2004.03705} {{Deep Learning Based Text Classification: A Comprehensive Review}}.
\newblock \emph{arXiv preprint arXiv:2004.03705}.

\bibitem[{{Mistral AI}(2025)}]{mistral}
{Mistral AI}. 2025.
\newblock {Medium is the new large.}
\newblock \url{https://mistral.ai/news/mistral-medium-3}.
\newblock Accessed: October 10, 2025.

\bibitem[{Mochtak et~al.(2025)Mochtak, Rupnik, Kuzman, and Ljube{\v{s}}i{\'c}}]{mochtak2025parlasent}
Michal Mochtak, Peter Rupnik, Taja Kuzman, and Nikola Ljube{\v{s}}i{\'c}. 2025.
\newblock \href {https://www.tandfonline.com/doi/full/10.1080/2474736X.2025.2508377} {Parlasent: mapping sentiment in political discourse with large language models}.
\newblock \emph{Political Research Exchange}, 7(1):2508377.

\bibitem[{Mochtak et~al.(2024)Mochtak, Rupnik, and Ljube{\v{s}}i{\'c}}]{mochtak2024parlasent}
Michal Mochtak, Peter Rupnik, and Nikola Ljube{\v{s}}i{\'c}. 2024.
\newblock \href {https://aclanthology.org/2024.lrec-main.1393/} {{The ParlaSent Multilingual Training Dataset for Sentiment Identification in Parliamentary Proceedings}}.
\newblock In \emph{Proceedings of the 2024 Joint International Conference on Computational Linguistics, Language Resources and Evaluation (LREC-COLING 2024)}, pages 16024--16036.

\bibitem[{Mochtak et~al.(2023)Mochtak, Rupnik, Meden, and Ljube{\v s}i{\'c}}]{parlasent-repository}
Michal Mochtak, Peter Rupnik, Katja Meden, and Nikola Ljube{\v s}i{\'c}. 2023.
\newblock \href {http://hdl.handle.net/11356/1868} {The multilingual sentiment dataset of parliamentary debates {ParlaSent} 1.0}.
\newblock Slovenian language resource repository {CLARIN}.{SI}: http://hdl.handle.net/11356/1868.

\bibitem[{OpenAI(2023)}]{openai}
OpenAI. 2023.
\newblock {ChatGPT General FAQ}.
\newblock \url{https://help.openai.com/en/articles/6783457-chatgpt-general-faq}.
\newblock Accessed: June 26, 2025.

\bibitem[{OpenAI(2024)}]{openai-gpt4o}
OpenAI. 2024.
\newblock {Hello GPT-4o}.
\newblock \url{https://openai.com/index/hello-gpt-4o/}.
\newblock Accessed September 11, 2024.

\bibitem[{OpenAI(2025)}]{gpt-5}
OpenAI. 2025.
\newblock {Introducing GPT-5}.
\newblock \url{https://openai.com/index/introducing-gpt-5/}.
\newblock Accessed: October 10, 2025.

\bibitem[{Orlikowski and Yates(1994)}]{orlikowski1994genre}
Wanda~J Orlikowski and JoAnne Yates. 1994.
\newblock \href {http://sullivanfiles.net/WID/assignments/discourse_field/genre_repertoire_orli_yates.pdf} {{Genre repertoire: The structuring of communicative practices in organizations}}.
\newblock \emph{Administrative science quarterly}, pages 541--574.

\bibitem[{Petukhova and Fachada(2023)}]{petukhova2023mn}
Alina Petukhova and Nuno Fachada. 2023.
\newblock \href {https://www.mdpi.com/2306-5729/8/5/74} {{MN-DS: A multilabeled news dataset for news articles hierarchical classification}}.
\newblock \emph{Data}, 8(5):74.

\bibitem[{{Qwen Team}(2024{\natexlab{a}})}]{team2024qwen2}
{Qwen Team}. 2024{\natexlab{a}}.
\newblock \href {https://arxiv.org/abs/2407.10671} {Qwen2 technical report}.
\newblock \emph{arXiv preprint arXiv:2407.10671}.

\bibitem[{{Qwen Team}(2024{\natexlab{b}})}]{qwen2.5}
{Qwen Team}. 2024{\natexlab{b}}.
\newblock \href {https://qwenlm.github.io/blog/qwen2.5/} {Qwen2.5: A party of foundation models}.
\newblock Accessed: June 26, 2025.

\bibitem[{Rupnik et~al.(2023)Rupnik, Ljubešić, and Mochtak}]{parlasent-model}
Peter Rupnik, Nikola Ljubešić, and Michal Mochtak. 2023.
\newblock \href {https://huggingface.co/classla/xlm-r-parlasent} {{Multilingual parliament sentiment regression model XLM-R-ParlaSent}}.
\newblock {Hugging Face}: https://doi.org/10.57967/hf/6718.

\bibitem[{Sun et~al.(2023)Sun, Li, Li, Wu, Guo, Zhang, and Wang}]{sun2023text}
Xiaofei Sun, Xiaoya Li, Jiwei Li, Fei Wu, Shangwei Guo, Tianwei Zhang, and Guoyin Wang. 2023.
\newblock \href {https://aclanthology.org/2023.findings-emnlp.603/} {{Text Classification via Large Language Models}}.
\newblock In \emph{Findings of the Association for Computational Linguistics: EMNLP 2023}, pages 8990--9005.

\bibitem[{Vaswani et~al.(2017)Vaswani, Shazeer, Parmar, Uszkoreit, Jones, Gomez, Kaiser, and Polosukhin}]{vaswani2017attention}
Ashish Vaswani, Noam Shazeer, Niki Parmar, Jakob Uszkoreit, Llion Jones, Aidan~N Gomez, {\L}ukasz Kaiser, and Illia Polosukhin. 2017.
\newblock \href {https://proceedings.neurips.cc/paper/2017/file/3f5ee243547dee91fbd053c1c4a845aa-Paper.pdf} {Attention is all you need}.
\newblock \emph{Advances in neural information processing systems}, 30.

\bibitem[{Yang et~al.(2025)Yang, Li, Yang, Zhang, Hui, Zheng, Yu, Gao, Huang, Lv et~al.}]{yang2025qwen3}
An~Yang, Anfeng Li, Baosong Yang, Beichen Zhang, Binyuan Hui, Bo~Zheng, Bowen Yu, Chang Gao, Chengen Huang, Chenxu Lv, et~al. 2025.
\newblock \href {https://arxiv.org/abs/2505.09388} {Qwen3 technical report}.
\newblock \emph{arXiv preprint arXiv:2505.09388}.

\bibitem[{Zhang et~al.(2022)Zhang, Ding, Jing, Dai, and Yin}]{zhang2022would}
Bowen Zhang, Daijun Ding, Liwen Jing, Genan Dai, and Nan Yin. 2022.
\newblock \href {https://arxiv.org/pdf/2212.14548} {{How would Stance Detection Techniques Evolve after the Launch of ChatGPT?}}
\newblock \emph{arXiv preprint arXiv:2212.14548}.

\bibitem[{Zhang et~al.(2025)Zhang, Wang, Li, Tiwari, and Qin}]{zhang2025pushing}
Yazhou Zhang, Mengyao Wang, Qiuchi Li, Prayag Tiwari, and Jing Qin. 2025.
\newblock \href {https://dl.acm.org/doi/abs/10.1145/3701716.3715528} {{Pushing the limit of LLM capacity for text classification}}.
\newblock In \emph{Companion Proceedings of the ACM on Web Conference 2025}, pages 1524--1528.

\bibitem[{Zhao et~al.(2024)Zhao, Chen, Zhang, and Yang}]{zhao2024advancing}
Hang Zhao, Qile~P Chen, Yijing~Barry Zhang, and Gang Yang. 2024.
\newblock \href {https://arxiv.org/abs/2412.08587} {{Advancing Single and Multi-task Text Classification through Large Language Model Fine-tuning}}.
\newblock \emph{arXiv preprint arXiv:2412.08587}.

\end{thebibliography}

% \bibliography{references}  %%% Remove comment to use the external .bib file (using bibtex).
%\printbibliography

\clearpage
\appendix

\section{Appendix}
\label{sec:appendix}

\subsection{Benchmarking Datasets}
\label{sec:app-test-datasets}

In this section, we provide additional information on the datasets used for benchmarking the models on sentiment identification, topic classification, and genre identification tasks in this study.

\paragraph{ParlaSent test datasets for sentiment classification in parliamentary speeches} include Croatian, Serbian, Bosnian, and English data from the multilingual sentiment dataset of parliamentary debates ParlaSent 1.0 (\citealp{mochtak2024parlasent}, \citealp{parlasent-repository}).\footnote{Available in the CLARIN.SI repository at \url{http://hdl.handle.net/11356/1585} and in the Hugging Face repository at \url{https://huggingface.co/datasets/classla/ParlaSent}.} The dataset comprises sentences that were randomly sampled from Croatian, Serbian, Bosnian and British parliamentary corpora and manually annotated with reported inter-annotator agreement ranging from 0.53 to 0.66 in Krippendorff's alpha \citep{krippendorff2018content}. The annotation involved a more granular six-level sentiment polarity scale that has been mapped to a three-level sentiment polarity scale which we use in our experiments: negative (0), neutral (1), and positive (2).

\paragraph{GINCO datasets for automatic genre identification} comprise the English EN-GINCO dataset \citep{kuzman2023automatic} and a multilingual X-GINCO dataset from the AGILE benchmark for Automatic Genre Identification.\footnote{\url{https://github.com/TajaKuzman/AGILE-Automatic-Genre-Identification-Benchmark}} The test instances were sampled from the enTenTen20 English web corpus \citep{jakubivcek2013tenten} and the MaCoCu multilingual web corpus collection \citep{banon2022macocu}. They were manually annotated by experts with a background in linguistics and computational linguistics who had experience with previous genre annotation campaigns \citep{kuzman-rupnik-ljubei:2022:LREC, kuzman2023automatic} where they reached an acceptable inter-annotator agreement of 0.71 in nominal Krippendorff's alpha \citep{krippendorff2018content}. While the X-GINCO dataset comprises numerous European languages, for the purposes of this study, we focus on three South Slavic languages: Croatian, Macedonian, and Slovenian. The test datasets use the X-GENRE annotation schema \citep{kuzman2023automatic} that includes the following genre labels: \textit{Information/Explanation}, \textit{News}, \textit{Instruction}, \textit{Opinion/Argumentation}, \textit{Forum}, \textit{Prose/Lyrical}, \textit{Legal} and \textit{Promotion}. While EN-GINCO and X-GINCO datasets have been annotated by the same annotator with the same schema, one should note that there are important differences between them in terms of their construction -- the English test dataset was sampled randomly from the web corpus, resulting in an unbalanced label distribution, while the X-GINCO datasets were curated with more deliberate interventions to ensure a balanced label distribution and a more controlled sampling process. Consequently, the X-GINCO datasets comprise fewer ambiguous instances and could be regarded as an easier test dataset.

\paragraph{IPTC News Topic test datasets} \citep{kuzman-iptc-classification} comprise Croatian and Slovenian news articles extracted from the MaCoCu-Genre web corpus collection \citep{macocu-genre} and manually annotated by one annotator. The reliability of the annotator was confirmed on a sample of data that was annotated by an additional annotator. The two annotators reached an acceptable inter-annotator agreement of 0.73 in nominal Krippendorff’s alpha \citep{krippendorff2018content}. Text instances are annotated with 17 topic labels from the top level of the IPTC NewsCodes Media Topic hierarchical schema, developed by the International Press Telecommunications Council (IPTC) \citep{iptcGroupsNewsCodes}. The datasets are more or less balanced by labels.

\paragraph{ParlaCAP test datasets} \citep{pungersek2026parlacap-paper} comprise parliamentary speeches in Bosnian, Croatian, English, and Serbian, sourced from the ParlaMint 4.1 dataset (\citealp{parlamint_41}; \citealp{erjavec2025parlamint}). These speeches were annotated by a single expert annotator using the 21 CAP categories from the official CAP schema \citep{baumgartner2019comparative}, along with an additional \textit{Other} label. The datasets are approximately balanced across labels. To assess the annotation quality, the Croatian dataset was independently annotated by two additional annotators. Inter-annotator agreement between the expert annotator and the others ranged from 0.62 to 0.68 in Krippendorff's alpha, which is around the threshold of 0.67 typically considered acceptable for annotation reliability \citep{krippendorff2018content}.

\subsection{Models}
\label{app:sec-models}

In the following subsections, we outline the models included in the evaluation -- the fine-tuned BERT-like classifiers (Section \ref{sec:bert-models}) and the open-weight and closed-source LLMs (Section \ref{sec:gpts}).

\subsubsection{Fine-Tuned BERT-like Models}
\label{sec:bert-models}

BERT (bidirectional encoder representations from transformers) deep neural models \citep{bert} have revolutionized the field of natural language processing (NLP), outperforming the non-neural methods across various NLP tasks. They have a more complex and computationally expensive architecture featuring transformers -- neural networks with self-attention mechanisms \citep{vaswani2017attention} -- that significantly improves the efficiency of training models on massive text data. Similarly to decoder-only transformer models, BERT models are pretrained on massive amounts of texts, possibly in multiple languages, which establishes their ability to encode the words and texts in high-dimensional vector spaces \citep{minaee2020deep} and enables their application even across languages in a zero-shot classification scenario. %The main difference between them is the method used for learning a textual representation: while autoregressive models predict a text sequence word by word based on the previous prediction, autoencoder models are trained by randomly masking some parts of the text sequence or corrupting the text sequence by replacing some of its parts \citep{minaee2020deep}.
To develop BERT-based classifiers, the pretrained models are trained, that is, fine-tuned, on a training dataset comprising text instances annotated with labels. In our study, we evaluate openly-accessible multilingual fine-tuned BERT-like models that have been already developed in recent related research. Namely, we evaluate the following models:
\begin{itemize}
    \item \textbf{IPTC News Topic classifier}\footnote{The IPTC News Topic classifier is available in the Hugging Face repository at \url{https://huggingface.co/classla/multilingual-IPTC-news-topic-classifier}.} \citep{kuzman-iptc-classification} is a multilingual fine-tuned BERT-like model for news topic classification according to the top-level IPTC NewsCodes schema \citep{iptcGroupsNewsCodes}. The model is based on the large-sized XLM-RoBERTa model \citep{conneau2020unsupervised} and was fine-tuned on 15,000 training text instances from the EMMediaTopic\footnote{The EMMediaTopic training dataset is available in the CLARIN.SI repository at \url{http://hdl.handle.net/11356/1991}.} dataset \citep{emmediatopic}. The training dataset contains news article instances in four languages: Catalan, Croatian, Greek, and Slovenian. The training dataset was annotated using an LLM that was shown to achieve annotation reliability comparable to that of human annotators \citep{kuzman-iptc-classification}. This approach is based on the novel methodology that uses the LLM teacher-student pipeline to develop BERT-like classifiers in the absence of manually-annotated training data.
    \item \textbf{XLM-R-ParlaSent} (\citealp{parlasent-model}; \citealp{mochtak2024parlasent}) is a domain-specific multilingual transformer model for sentiment identification in parliamentary texts. It is based on the XLM-R-parla pretrained model \citep{xlm-r-parla} that was developed by additionally pretraining the large-sized XLM-RoBERTa model \citep{conneau2020unsupervised} on 1.72 billion words from parliamentary proceedings in 30 European languages.
    To develop the XLM-R-ParlaSent model,\footnote{The XLM-R-ParlaSent model is accessible in the Hugging Face repository at \url{https://huggingface.co/classla/xlm-r-parlasent}.} the pretrained XLM-R-Parla model was fine-tuned on the ParlaSent sentiment training dataset (\citealp{mochtak2024parlasent}; \citealp{parlasent-repository}) in seven European languages (Bosnian, Croatian, Czech, English, Serbian, Slovak, and Slovenian). The training dataset\footnote{The ParlaSent training and test datasets are freely available in the CLARIN.SI repository at \url{http://hdl.handle.net/11356/1868}.} comprises 13,000 instances sampled from parliamentary proceedings and manually annotated with sentiment labels. %sampled from parliamentary proceedings of seven European countries (Bosnia and Herzegovina, Croatia, Czech Republic, Serbia, Slovakia, Slovenia, and United Kingdom). The training dataset consists of 2,600 instances per each of the five parliaments, 13,000 instances in total. The XLM-R-Parla model is used as the base model instead of the XLM-RoBERTa, as previous experiments have shown that the additionally pretrained model outperforms the XLM-RoBERTa model \citep{mochtak2024parlasent}.
    \item \textbf{ParlaCAP classifier}\footnote{The ParlaCAP topic classifier is available in the Hugging Face repository at \url{https://huggingface.co/classla/ParlaCAP-Topic-Classifier}.} (\citealp{parlacap_model}; \citealp{pungersek2026parlacap-paper}) is a domain-specific multilingual transformer model for topic classification in parliamentary texts based on the CAP schema \citep{baumgartner2019comparative}. As the XLM-R-ParlaSent model, this model is based on the XLM-R-parla pretrained model (\citealp{xlm-r-parla}; \citealp{mochtak2024parlasent}). The XLM-R-parla model was then fine-tuned on the ParlaCAP-train dataset\footnote{The ParlaCAP-train training dataset is available in the CLARIN.SI repository at \url{http://hdl.handle.net/11356/2093}.} (\citealp{parlacap-train}; \citealp{pungersek2026parlacap-paper}). The training dataset comprises around 30 thousand speeches from parliamentary debates from the ParlaMint 4.1 parliamentary datasets (\citealp{parlamint_41}; \citealp{erjavec2025parlamint}) in 29 European languages. The training dataset was annotated with the CAP categories by a GPT-4o \citep{openai-gpt4o} model used in a zero-shot prompting fashion, following the LLM teacher-student framework \citep{kuzman-iptc-classification}. Based on the inter-annotator agreement, calculated on a sample that was annotated by three human annotators and the LLM annotator, the agreement between the LLM and the human annotators was comparable to the agreement between the human annotators themselves. This indicates that the LLM performs as reliably as human annotators on this task, supporting its use for annotating the training data.
    \item \textbf{X-GENRE classifier} (\citealp{kuzman2023automatic}; \citealp{x-genre-repository}) is a multilingual fine-tuned BERT-like model for automatic genre identification.\footnote{The X-GENRE classifier is freely available in the Hugging Face repository at \url{https://doi.org/10.57967/hf/0927} and the CLARIN.SI repository at \url{http://hdl.handle.net/11356/1961}.} The model is based on the base-sized XLM-RoBERTa model \citep{conneau2020unsupervised} and was fine-tuned on the training split of the X-GENRE dataset \citep{x-genre-dataset}, which contains 1,772 text instances in Slovenian and English, manually-annotated with genre labels from the X-GENRE schema \citep{kuzman2023automatic}.
\end{itemize}

\subsubsection{Instruction-Tuned Large Language Models}
\label{sec:gpts}

As the BERT models, decoder-only large language models are based on a transformer deep neural architecture and are pretrained on massive text collections. However, while the development of fine-tuned BERT-like classifiers necessitates large amounts of annotated training data, recent advances in the field have shown that the instruction-tuned LLMs are capable of text classification in a zero-shot or few-shot prompting setups which does not require any training data. We assess the performance of the following large language models:
\begin{itemize}
    \item \textbf{OpenAI models}, namely the GPT-3.5-Turbo (\texttt{gpt-3.5-turbo-0125}) \citep{openai}, GPT-4o (\texttt{gpt-4o-2024-08-06}) \citep{openai-gpt4o} and the GPT-5 (\texttt{gpt-5-2025-08-07}) \citep{gpt-5}. These closed-source instruction-tuned LLMs were developed by OpenAI. OpenAI states that the models are trained on large multilingual web corpora, however, specific details about the training data, procedures, and architecture are not publicly known.
    \item \textbf{Gemini 2.5 Flash model} \citep{comanici2025gemini} is a closed-source multilingual and multimodal instruction-tuned LLM by Google DeepMind. The model is reported to be pretrained on over 400 languages \citep{comanici2025gemini}, however, details on the language coverage are not available.
    \item \textbf{Mistral Medium 3.1 model} (\texttt{mistral-medium-2508}) \citep{mistral} is a closed-source multimodal instruction-tuned model by Mistral AI. Available details on the model architecture and language coverage are very limited.
    \item \textbf{LLaMA 3.3 model}\footnote{\url{https://ollama.com/library/llama3.3}} \citep{llama33modelcard} is an open-weight instruction-tuned multilingual LLM, developed by Meta, with 70 billion parameters. The model was pretrained on a web text collection in various languages, however, it is reported to support only 8 languages, namely, English, German, French, Italian, Portuguese, Hindi, Spanish, and Thai.
    \item \textbf{Gemma 3 model}\footnote{\url{https://ollama.com/library/gemma3}} \citep{gemma3} is an open-weight multilingual instruction-tuned LLM, developed by Google DeepMind. The model was pretrained on multimodal data with large quantities of multilingual texts and is reported to support over 140 languages. We use the model in 27 billion parameter size.
    \item \textbf{DeepSeek-R1-Distill}\footnote{\url{https://ollama.com/library/deepseek-r1:14b}} \citep{guo2025deepseek} is an open-weight reasoning LLM, developed by DeepSeek AI. We use the distilled model in 14 billion parameter size, namely the \texttt{DeepSeek-R1-Distill-Qwen-14B} model. The model is based on the Qwen 2.5 model \citep{qwen2.5,team2024qwen2} that was fine-tuned using a dataset curated with the DeepSeek-R1 reasoning model. The Qwen 2.5 model provides multilingual support for over 29 languages, including Chinese, English, French, Spanish, Portuguese, German, Italian, Russian, Japanese, Korean, Vietnamese, Thai, and Arabic.
    \item \textbf{Qwen 3}\footnote{\url{https://ollama.com/library/qwen3}} (\texttt{Qwen3-2504}) \citep{yang2025qwen3} is an open-weight LLM, developed by Alibaba Cloud. We use the model with the 32 billion parameter size, namely, the \texttt{qwen3:32b} model. The model is said to support over 100 languages and dialects \citep{yang2025qwen3}.
\end{itemize}

Open-weight models were installed locally and executed via the Ollama API service \citep{maric_local_llms}. We use the quantized versions of the models as they are available through the Ollama library.\footnote{\url{https://ollama.com/library}} OpenAI models are used through the chat completion endpoint via the OpenAI API, whereas other closed-source models were accessed through the OpenRouter platform\footnote{\url{https://openrouter.ai/}} that provides a unified API access to various closed-source models.

\begin{figure*}[htbp]
    \centering

    \begin{subfigure}[b]{0.45\textwidth}
        \centering
        \includegraphics[width=\textwidth]{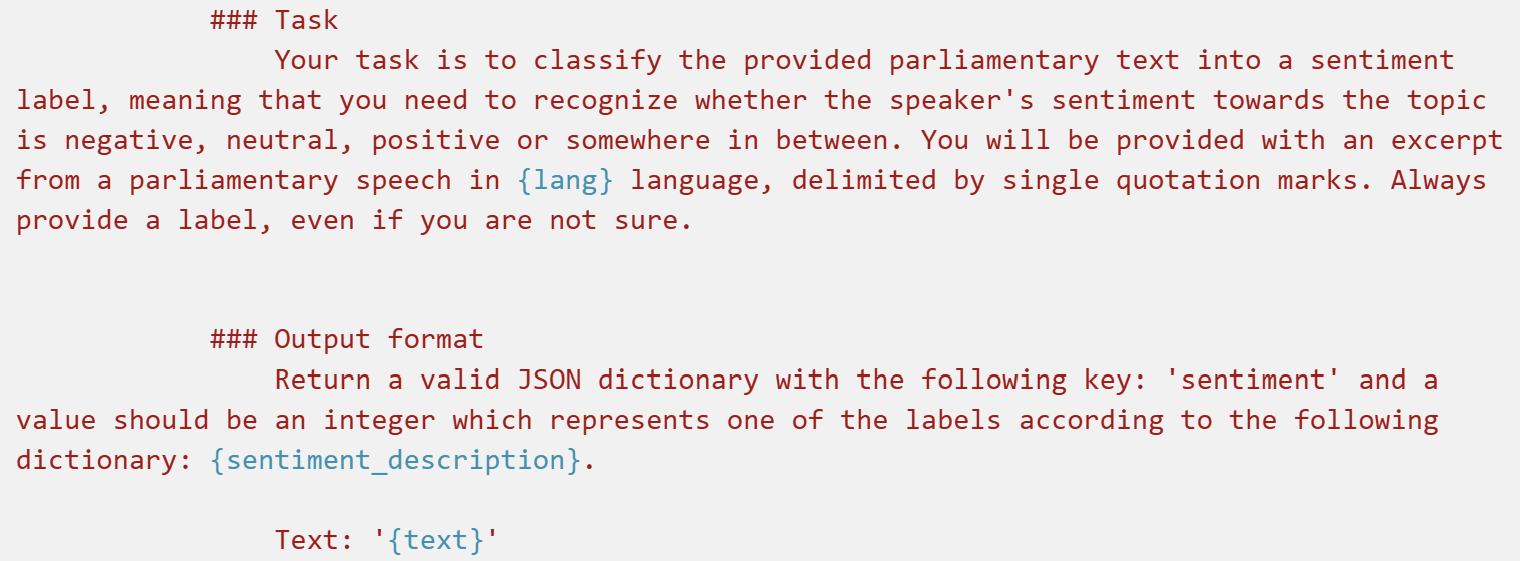}
        \caption{Sentiment classification.}
        \label{fig:sentiment-prompt}
    \end{subfigure}
    \hfill
    \begin{subfigure}[b]{0.45\textwidth}
        \centering
        \includegraphics[width=\textwidth]{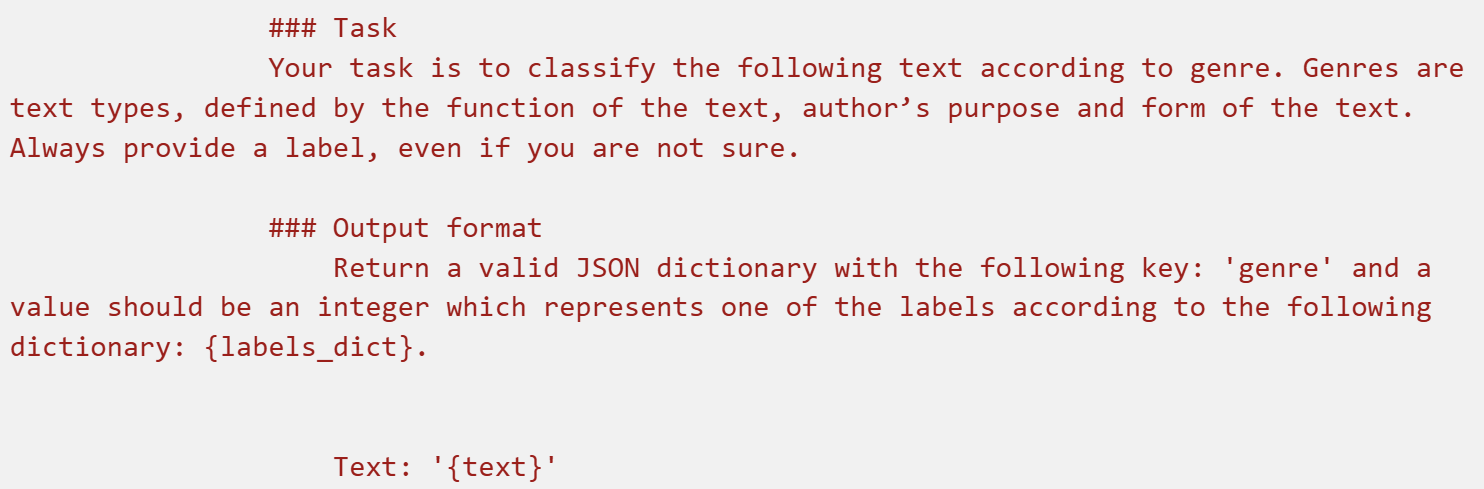}
        \caption{Automatic genre identification.}
        \label{fig:genre-prompt}
    \end{subfigure}

    \vskip\baselineskip 

    % Second row
    \begin{subfigure}[b]{0.45\textwidth}
        \centering
        \includegraphics[width=\textwidth]{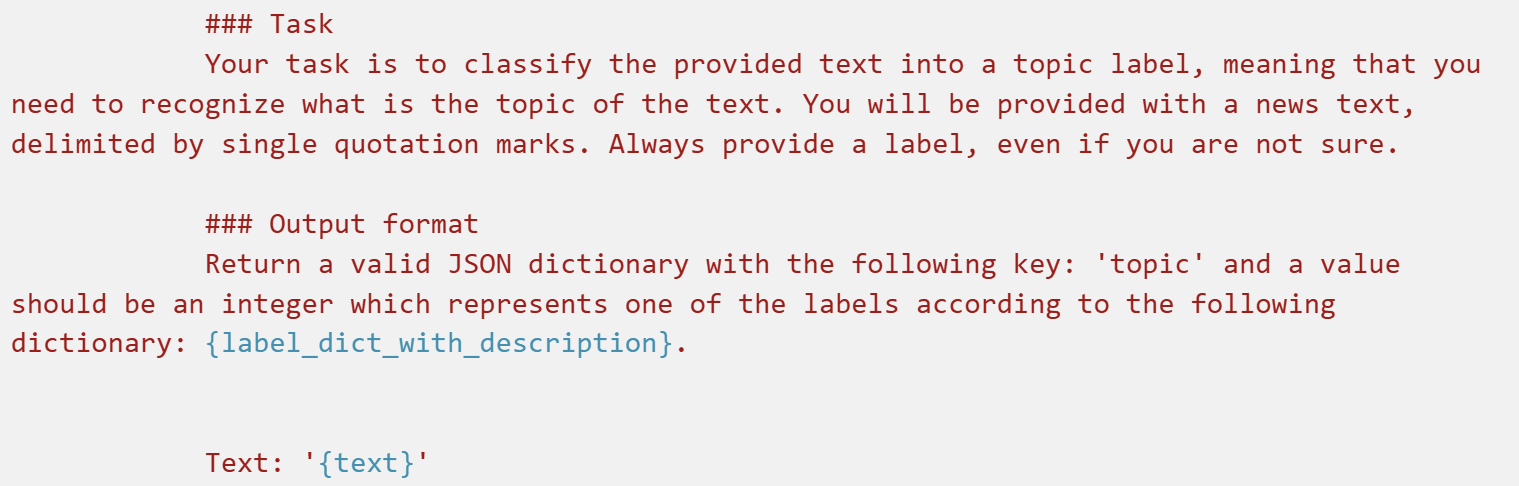}
        \caption{News topic classification.}
        \label{fig:topic-prompt}
    \end{subfigure}
    \hfill
    \begin{subfigure}[b]{0.45\textwidth}
        \centering
        \includegraphics[width=\textwidth]{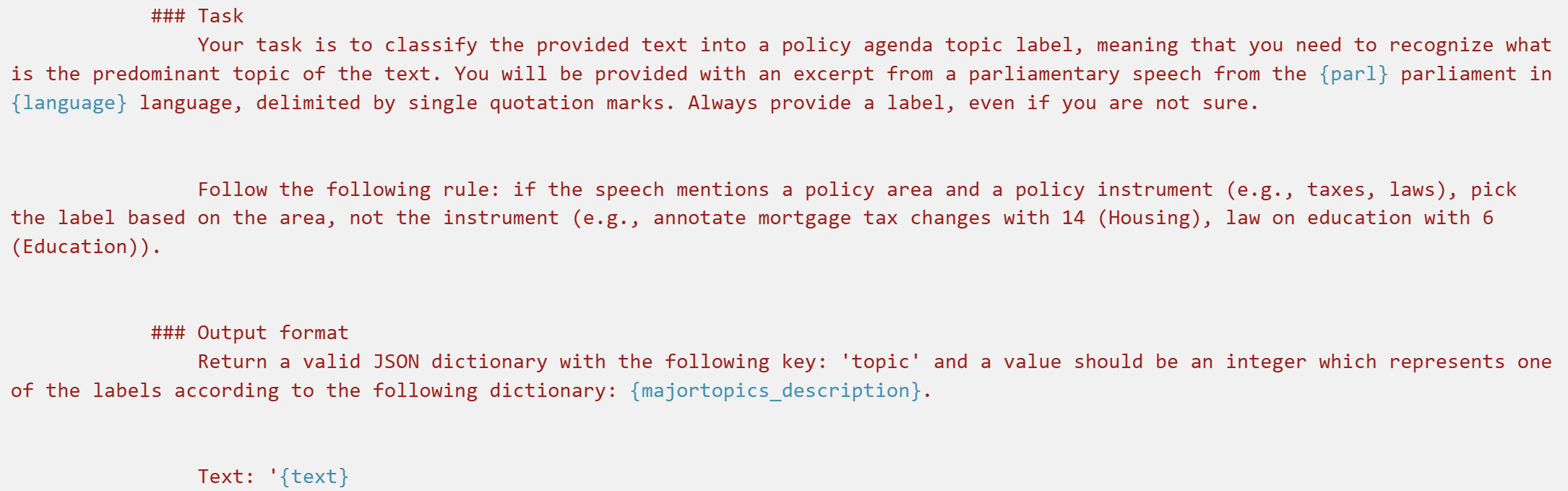}
        \caption{Parliamentary topic classification.}
        \label{fig:parlacap-prompt}
    \end{subfigure}
    
    \caption{The prompts that are provided to the LLMs for the sentiment identification task (Figure \ref{fig:sentiment-prompt}), automatic genre identification (Figure \ref{fig:genre-prompt}), and topic classification on news (Figure \ref{fig:topic-prompt}) and parliamentary speeches (Figure \ref{fig:parlacap-prompt}). The prompts comprise the description of the task and labels with a short description.}
    \label{fig:prompts}
\end{figure*}

To prevent any bias, all models were used with their default parameters. The only parameter that we defined is the temperature which we set to 0 to ensure a more deterministic behaviour of the models. The same prompts were used for all open-weight and closed-source models. In Figure \ref{fig:prompts}, we provide prompts that were provided to the LLMs for zero-shot text classification, namely for sentiment classification (Figure \ref{fig:sentiment-prompt}), automatic genre identification (Figure \ref{fig:genre-prompt}), news topic classification (Figure \ref{fig:topic-prompt}) and topic classification in parliamentary speeches (Figure \ref{fig:parlacap-prompt}). For more details on the setups used to apply fine-tuned BERT-like models and instruction-tuned LLMs to the test datasets, refer to the code published on GitHub.\footnote{\url{https://github.com/TajaKuzman/Benchmarking-Text-Classification-on-South-Slavic}}

\end{document}